\documentclass[lettersize,journal]{IEEEtran}
\usepackage{amsmath,amsfonts}
\usepackage{array}
\usepackage[caption=false,font=normalsize,labelfont=sf,textfont=sf]{subfig}
\usepackage{textcomp}
\usepackage{stfloats}
\usepackage{url}
\usepackage{verbatim}
\usepackage{graphicx}
\usepackage{cite}
\usepackage{multirow} 
\usepackage{pifont} 
\usepackage{colortbl} 
\usepackage{xcolor}
\usepackage{algorithm}
\usepackage{algpseudocode}

\usepackage{hyperref}
\hypersetup{
	colorlinks=true,
	linkcolor=cyan,
	filecolor=blue,      
	urlcolor=black,
	citecolor=green,
}

\begin{document}

\title{CurbNet: Curb Detection Framework Based on LiDAR Point Cloud Segmentation}


\author{Guoyang Zhao, Fulong Ma, Weiqing Qi, Yuxuan Liu, Ming Liu, and Jun Ma, \textit{Senior Member, IEEE}

\thanks{This work was supported in part by the National Natural Science Foundation of China under Grant 62303390; in part by the Guangdong Provincial Key Lab of Integrated Communication, Sensing and Computation for Ubiquitous Internet of Things under Grant 2023B1212010007; in part by the Guangzhou-HKUST(GZ) Joint Funding Scheme under Grant 2024A03J0618. \textit{(Corresponding author: Jun Ma.)}}
\thanks{Guoyang Zhao, Fulong Ma, Weiqing Qi, and Ming Liu are with the Robotics and Autonomous Systems Thrust, The Hong Kong University of Science and Technology (Guangzhou), Guangzhou 511453, China
(e-mail: gzhao492@connect.hkust-gz.edu.cn; fmaaf@connect.hkust-gz.edu.cn; wqiad@connect.hkust-gz.edu.cn; eelium@hkust-gz.edu.cn)}
\thanks{Yuxuan Liu is with the Department of Electronic and Computer Engineering, The Hong Kong University of Science and Technology, Hong Kong SAR, China
(e-mail: yliuhb@connect.ust.hk)}
\thanks{Jun Ma is with the Robotics and Autonomous Systems Thrust, The Hong Kong University of Science and Technology (Guangzhou), Guangzhou 511453, China, and also with the Division of Emerging Interdisciplinary Areas, The Hong Kong University of Science and Technology, Hong Kong SAR, China (e-mail: jun.ma@ust.hk).} 
}
\markboth{IEEE Transactions on Intelligent Transportation Systems}%
{Shell \MakeLowercase{\textit{et al.}}: A Sample Article Using IEEEtran.cls for IEEE Journals}


\maketitle

\begin{abstract}
Curb detection is a crucial function in intelligent driving, essential for determining drivable areas on the road. However, the complexity of road environments makes curb detection challenging. This paper introduces CurbNet, a novel framework for curb detection utilizing point cloud segmentation. To address the lack of comprehensive curb datasets with 3D annotations, we have developed the 3D-Curb dataset based on SemanticKITTI, currently the largest and most diverse collection of curb point clouds.
Recognizing that the primary characteristic of curbs is height variation, our approach leverages spatially rich 3D point clouds for training. To tackle the challenges posed by the uneven distribution of curb features on the xy-plane and their dependence on high-frequency features along the z-axis, we introduce the Multi-Scale and Channel Attention (MSCA) module, a customized solution designed to optimize detection performance. Additionally, we propose an adaptive weighted loss function group specifically formulated to counteract the imbalance in the distribution of curb point clouds relative to other categories.
Extensive experiments conducted on 2 major datasets demonstrate that our method surpasses existing benchmarks set by leading curb detection and point cloud segmentation models. 
Through the post-processing refinement of the detection results, we have significantly reduced noise in curb detection, thereby improving precision by 4.5 points. 
Similarly, our tolerance experiments also achieve state-of-the-art results.
Furthermore, real-world experiments and dataset analyses mutually validate each other, reinforcing CurbNet's superior detection capability and robust generalizability. The project website is available at: \url{https://github.com/guoyangzhao/CurbNet/}.

\end{abstract}


\section{Introduction}

Autonomous vehicles fundamentally depend on analyzing data from onboard sensors to understand their surrounding environment, a cornerstone for safe driving \cite{luettel2012autonomous,ma2023every}. In this context, road boundary detection is a crucial aspect of perception, delineating road and non-road areas \cite{xu2021icurb}. This distinction is vital for the positioning, planning, and decision-making of self-driving cars, especially under conditions where GPS signals are obscured by trees and buildings. In such urban road environments, curbs serve as a key and effective feature for vehicle localization \cite{suhr2016sensor,hata2017monte}. Despite their importance, curbs, typically linear, thin, and long, pose significant challenges for detection in complex road environments \cite{romero2021road}.

\begin{figure}[t]
    \centering
    \includegraphics[width=0.5\textwidth]{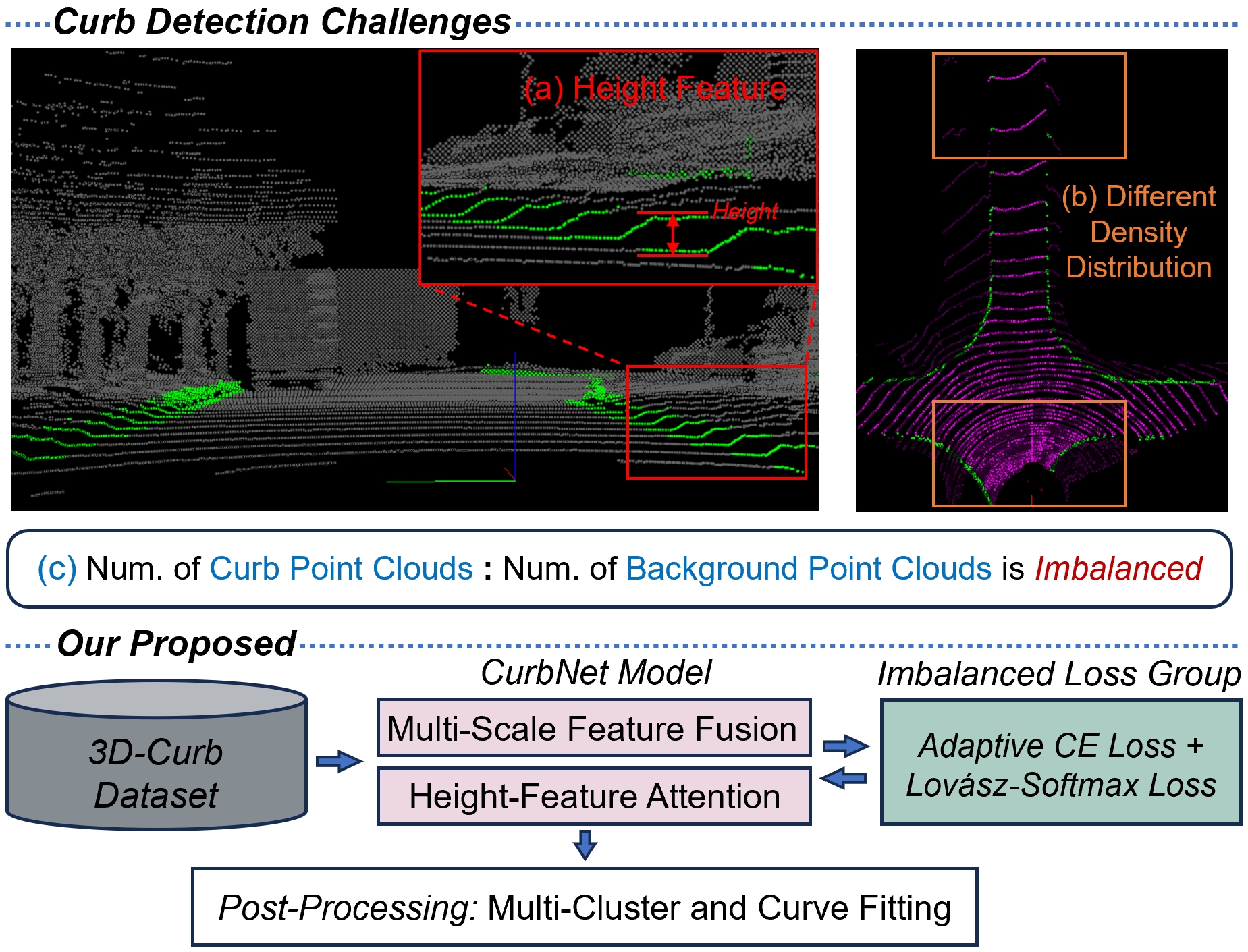}
    \vspace{-15pt}
    \caption{\textbf{Curb detection challenges and our proposed method.} Three main challenges of curb detection are shown (a) height feature extraction (b) different density distribution of point clouds (c) Curb point cloud quantity proportion imbalance. Solution: First propose a 3D-Curb dataset. The MSCA module is designed for multi-scale spatial feature fusion and height feature extraction. The loss group is proposed to solve the imbalance problem. Finally, we use post-processing to further improve performance.}
    \label{cover-figure}
    \vspace{-15pt}
\end{figure}

Research in curb detection varies based on sensor technology, broadly categorized into vision-based and LiDAR-based methods \cite{hillel2014recent}. This includes the use of monocular cameras, stereo vision, and combinations of 2D and 3D LiDAR sensors \cite{sun20193d}. Vision-based methods can provide rich contextual information and achieve effective detection results. However, camera performance is greatly influenced by light and weather conditions and fails to provide accurate depth information directly, which is a critical need for autonomous driving applications \cite{wei2022row, zhao2024fisheyedepth}. Furthermore, the subtle color differences between regular road surfaces and curbs make it difficult for cameras to accurately detect curbs. In contrast, LiDAR sensors demonstrate robustness under various weather and lighting conditions and offer precise distance measurements \cite{ma2022automatic}. Recently, 3D LiDAR has become one of the most important sensors for perceiving 3D environments in autonomous vehicles \cite{demir2017adaptive}.

Curb detection using LiDAR can be classified into manual feature methods and learning based methods \cite{horvath2021real}. Manual feature design methods \cite{wang2020speed,chen2015velodyne} typically analyze geometric relationships such as height and angle changes between adjacent points, given the difference between drivable roads and curbs \cite{hata2014robust,zhou2012mapping}. Most curb detection methods follow a sequential point cloud processing procedure \cite{xu2016road,zhang2018road}, including stages of candidate region extraction, manual feature setting and clustering, and post-processing for fitting estimation. These methods, due to their interpretability in terms of safety, are widely used in curb detection for autonomous driving systems. However, the reliability of these manually designed rule-based processes is limited in practical applications, as errors in early detection stages can severely impact subsequent recognition performance. Additionally, they rely on manual feature design heavily, necessitating extensive parameter adjustments for various scenarios such as straight roads, curved roads, and different types of intersections, resulting in low practical efficiency and generalization \cite{hata2014robust,qin2012curb}.

With the breakthroughs in deep learning applications in perception, its capabilities in automatic feature extraction and learning have significantly reduced the tedium of manual feature design and parameter adjustment \cite{qi2024clrkdnet, zhao2021real}. Particularly, in addressing different road scenarios, it has substantially enhanced model recognition performance and robustness \cite{xu2021cp}. Recently, some researchers have explored curb detection using CNN-based methods, projecting 3D LiDAR point cloud data into 2D images from a Bird's Eye View (BEV) perspective, followed by processing these images with CNN models to detect curbs \cite{jung2021uncertainty,suleymanov2019online}. However, this direct projection of 3D LiDAR data can lead to the loss of essential spatial structural information, particularly the crucial height difference information for curb detection \cite{gao2023lcdet}.

Considering current research in curb detection and the physical properties of 3D LiDAR, we have identified several challenges that need addressing, as shown in Fig. ~\ref{cover-figure}: (a) The primary distinguishing feature of curbs is the subtle height variation from the road surface \cite{wang2020speed}, a challenging feature for models to accurately learn. (b) Curb point clouds from LiDAR scanning show significant distribution differences across various distances \cite{apellaniz2023lidar}. (c) Curbs occupy only a small portion of LiDAR point clouds, appearing as long and narrow lines, making effective model training difficult \cite{gao2023lcdet}.

We first propose the 3D-Curb dataset, which contains 3D annotated point cloud data with a wide range of road scenes. 
Unlike most existing algorithms that project 3D point clouds onto 2D images, we extract features directly from point clouds, preserving crucial 3D spatial information. 
To address challenges (a) and (b), we propose a Multi-Scale and Channel Attention (MSCA) module. This module employs a multi-scale fusion stage to mitigate the uneven distribution of curb point clouds and a channel attention stage to dynamically capture height variations along the z-axis feature.
For challenge (c), which pertains to the severe imbalance in point cloud quantities among different categories, we introduce a novel loss combination. This includes an adaptive cross-entropy (ACE) loss and an IoU-focused loss, designed to handle the imbalanced data distribution effectively. Moreover, recognizing the inherent sparsity of curb point clouds, we augment the training process by adding curb-related categories (such as roads and sidewalks) to assist in model learning.
Finally, to further improve the precision of curb detection, we propose a post-processing scheme of multi-cluster and then fitting, which effectively removes the noise around the curb detection results.

Our primary contributions are summarized as follows:
\begin{itemize}
\item[1)] We introduce a comprehensive 3D-Curb point cloud dataset based on SemanticKITTI, which is the largest and most diverse currently available to our knowledge.
\item[2)] We propose a novel Multi-Scale and Channel Attention (MSCA) module and an imbalance loss strategy, tailored to the distribution characteristics of curb point clouds.
\item[3)] We develop a multi-cluster fitting post-processing approach to further enhance detection performance.
\item[4)] Our method achieves state-of-the-art detection results in complex, large-scale intersection scenarios.

\end{itemize}

\section{Related Works}
\subsection{Manual Feature Extraction Methods}
In the early development of LiDAR technology, researchers designed mathematical functions to manually extract curb features by understanding the principles of LiDAR scanning. This process primarily involves stages of feature point extraction, feature point classification and filtering, and curb curve fitting and estimation. In \cite{zai20173}, feature points are extracted through image segmentation and energy minimization, followed by the application of principal curves and surfaces methods in \cite{ozertem2011locally} for fitting detected curbs. Studies like \cite{wang2020speed,chen2015velodyne,jie2022efficient,yao2012road} utilize the horizontal and vertical continuity of point clouds, employing angle and height thresholds for curb feature extraction and using Gaussian Process Regression (GPR) and Random Sample Consensus for curb curve fitting. \cite{hata2014robust,chen2015velodyne} integrate the generalized curvature method from LOAM \cite{zhang2014loam} into curb detection, refining the process with GPR. \cite{hata2015feature,hata2014robust} detect curbs by analyzing ring compression in dense 3D LIDAR data, employing false positive filters and least-squares regression filters based on height values, respectively. \cite{zhang2018road,yang2013semi} propose sliding-beam segmentation and sliding-window detection methods by analyzing individual LiDAR scan lines, focusing on specific curb detection in each frame.

However, these manual feature extraction and sequential processing methods are inefficient. Not only is feature creation laborious and requires specialized knowledge, but early-stage erroneous selections can impact later detection phases, making them inadequate for complex road scenes and diverse curb shapes \cite{bai2022build}. In this context, deep learning methods can effectively solve the issues of feature extraction and generalization.

\subsection{Deep Learning Methods}
With the advancement of deep learning, some researchers have begun using CNNs to detect curbs in LiDAR point clouds, achieving notable performance. A common characteristic of deep learning-based curb detection methods is the transformation of input 3D point cloud data into 2D view images or voxelization. \cite{liang2019convolutional} uses camera images, LiDAR, and elevation gradients of LiDAR as inputs, employing convolutional recurrent networks to extract road boundaries in 2D BEV images to construct semantic maps. \cite{suleymanov2019online} projects motion-accumulated 3D point cloud data onto 2D BEV images, initially detecting visible road edges through a U-Net network \cite{ronneberger2015u}, followed by predicting obscured road boundaries using multi-layer convolutional networks with expanded receptive fields \cite{pan2018spatial}. Similarly, \cite{jung2021uncertainty} proposes a two-stage curb detection framework, initially employing U-Net for visible curb detection, then incorporating uncertainty quantification to improve detection performance in obscured areas.

Compared to traditional methods, these approaches demonstrate robustness in various driving environments and reduce the burden of manual parameter tuning. However, converting point clouds to 2D images results in significant loss of 3D information, especially the height difference features crucial for distinguishing curbs from other categories. \cite{kukolj2023road} explores voxelizing the raw 3D point cloud as a superimage input to the model, using CNNs to detect road edges and lane markings, but its recognition accuracy remains low. LCDeT \cite{gao2023lcdet} employs a Transformer model for curb detection in voxelized point clouds, introducing dual attention mechanisms in both temporal and spatial dimensions to ensure detection stability and accuracy. Yet, this method relies on complex model structures and high-resolution LiDAR sensors.

Some general point cloud segmentation models \cite{cciccek20163d, zhou2020cylinder3d, hou2022point} have achieved accurate 3D point cloud detection using relatively simple model structures. Among them, 3D U-Net \cite{cciccek20163d} is a popular model for 3D volumetric data processing, adapted for point cloud segmentation by learning the spatial hierarchies of features. Cylinder3D \cite{zhou2020cylinder3d} is a voxel-based method specifically designed to handle large-scale point clouds by projecting them into cylindrical grids, enabling more efficient feature extraction in 3D space. PVKD \cite{hou2022point} is an improvement of Cylinder3D based on knowledge distillation, focusing on transferring knowledge from a large teacher model to a smaller student network to enhance segmentation performance. However, these models are not specifically designed to extract curb features, which are crucial for curb detection tasks.

Our proposed method uses original point cloud as input, preserving more 3D feature information. Addressing the uneven distribution of curb point cloud feature and the easy loss of curb height difference information, we introduce a multi-scale and channel attention mechanisms to enhance performance.

\subsection{Curb Detection Datasets}
There is a significant amount of research in LiDAR-based curb detection; however, high-quality datasets with 3D annotations are scarce. Major automotive datasets such as NuScenes \cite{caesar2020nuscenes}, KITTI \cite{geiger2013vision}, and SemanticKITTI \cite{behley2019semantickitti} do not include curb annotations. The robustness of deep learning methods is closely related to the volume of data collected under various environmental conditions. These factors have somewhat hindered the application of deep learning methods in this task. \cite{zhang2018road} created a public dataset for curb detection, comprising 200 scans collected across five different scenes. \cite{jung2021uncertainty} developed a public dataset for curb detection, including 5200 scans with BEV labels, collected from urban areas. LCDeT \cite{gao2023lcdet} introduced a curb dataset for 128-line LiDAR, containing 6200 frames of point clouds from urban road scenes, both during daytime and nighttime. \cite{apellaniz2023lidar} proposed a method for 3D curb detection and annotation in LiDAR point clouds, effectively reducing manual annotation time by 50\%.

The above-mentioned curb point cloud datasets only label the curb category, which cannot be directly used in autonomous driving scenarios, as recognition of other categories such as vehicles and roads is also necessary. Similarly, other categories surrounding the curb, like roads and sidewalks, can also assist the model in more accurately learning curb features. Based on the existing large SemanticKITTI dataset, we added annotations for the curb category, thereby covering a richer array of real road scenes, totaling up to 7100 frames of point cloud data. To our knowledge, this is currently the largest and most comprehensive curb point cloud dataset with annotations relevant to autonomous vehicles (AV). Table~\ref{dataset} illustrates the related LiDAR datasets for curb detection.

\begin{figure}[t]
    \centering
    \includegraphics[width=0.5\textwidth]{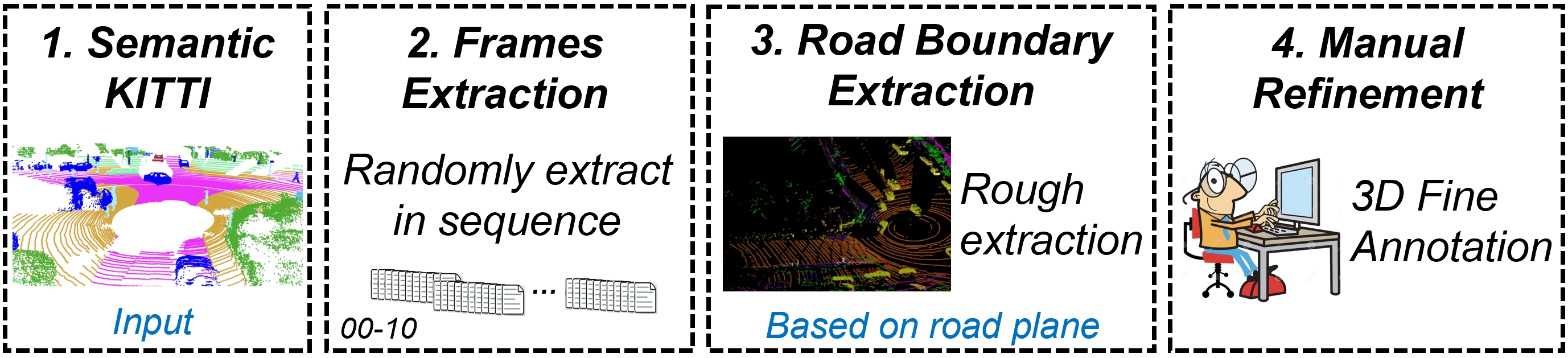}
    \vspace{-15pt}
    \caption{\textbf{3D-Curb dataset construction process.} Mainly developed based on the standard SemanticKITTI dataset.}
    \label{dataset-figure}
    \vspace{-5pt}
\end{figure}

\begin{table}[t!]
\renewcommand\arraystretch{1.5}
\caption{COMPARISON OF RELATED CURB DATASETS.}
\centering
\setlength{\tabcolsep}{1.5mm}
\footnotesize
\begin{tabular}{cccccc}
\hline
Dataset & LiDAR & Public & Frames & 3D labeling & Categories\\
\hline
Zhang\cite{zhang2018road} & 32-line & Yes & 200 & No & 1\\
Liang\cite{liang2019convolutional} & - & No & - & No & 1\\
Suleymanov\cite{suleymanov2019online} & - & No & - & No & 1\\
Uncertainty\cite{jung2021uncertainty} & 32-line & Yes & 5224 & No & 2\\
NRS\cite{gao2023lcdet} & \textbf{128-line} & Yes & 6220 & No & 2\\
3D-Curb (ours) & 64-line & \textbf{Yes} & \textbf{7100} & \textbf{Yes} & \textbf{29}\\
\hline
\end{tabular}
\label{dataset}
\end{table}

\section{Methodology}

\subsection{3D-Curb Dataset Construction} 

Compared to other autonomous driving scenarios, there is a significant lack of relevant curb point cloud datasets, especially those with 3D  annotations. Building on the large-scale open-source SemanticKITTI dataset \cite{behley2019semantickitti}, we have developed and introduced the 3D-Curb dataset. This dataset retains the original 28 semantic categories while adding a new curb category. It was collected using a Velodyne HDL-64E LiDAR, providing comprehensive views of various street scenes as a general-purpose autonomous driving dataset.

The construction process of our dataset is illustrated in Fig.~\ref{dataset-figure}. Due to the high-frequency data acquisition of the SemanticKITTI dataset, the high similarity between frames can lead to model overfitting and significantly increase the annotation workload. Therefore, we randomly selected 7100 representative frames from sequences 00-10. We utilized the high-quality road annotations provided by the SemanticKITTI dataset and applied the Ground Plane Fitting method proposed by \cite{zermas2017fast} to extract the boundaries within the road labels. We then extended these parameters to obtain the curb labels. However, due to static/dynamic obstacles and other occlusions, these automatically generated curb annotations contained many inaccuracies. Thus, we manually refined the curb labels in the BEV perspective to ensure high-quality annotations. The 3D-Curb dataset focuses on curb annotations in the forward direction of vehicle travel, with an average range of 40.43 meters along the forward y-axis. To accentuate the curb areas, the lateral x-axis range is set to 1.3 times the road width.

To the best of our knowledge, this is the largest curb point cloud dataset to date and the only one with 3D annotations. Table~\ref{dataset} compares our dataset with other related curb datasets.

\begin{figure*}[t!]
    \centering
    \includegraphics[width=.99\textwidth]{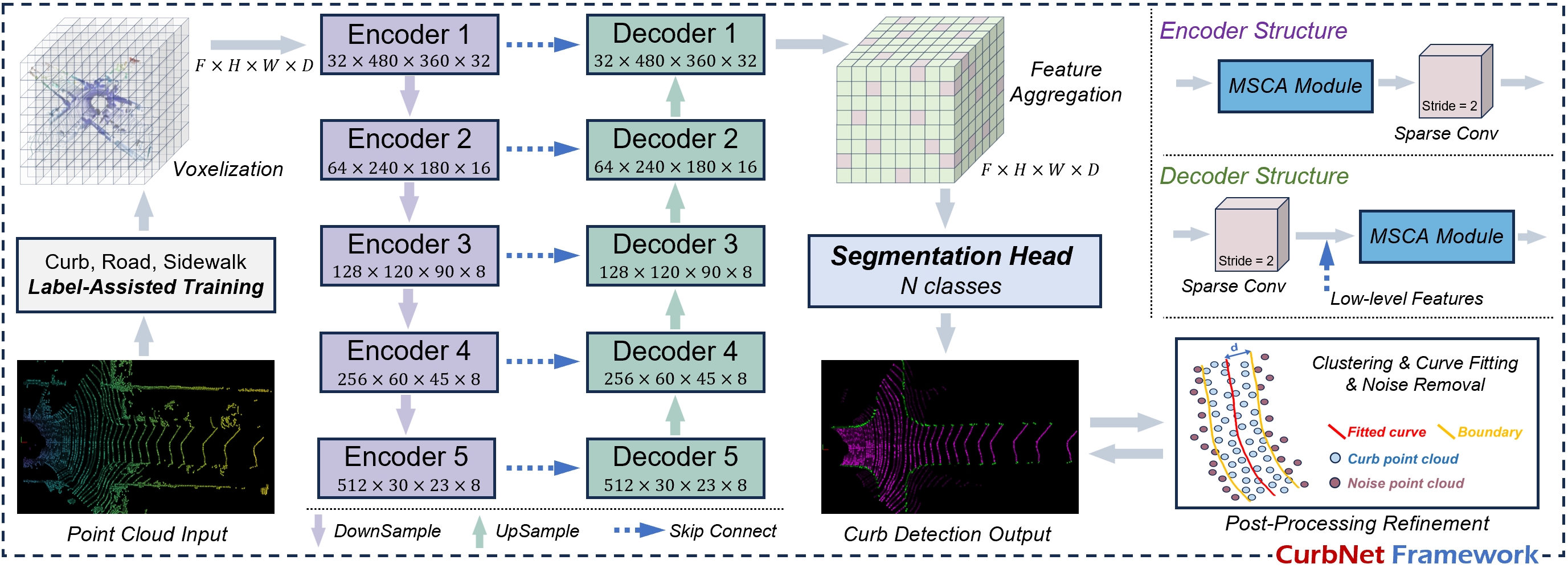}
    \vspace{-10pt}
    \caption{\textbf{Overview of proposed CurbNet framework.} From left to right, first is point cloud data input and voxelization.  Then there is a 5-layer deep encoder-decoder structure.  Next comes the feature aggregation and segmentation head.  Finally, the post-processing refinement of the detection results.}
    \label{framework}
    \vspace{-13pt}
\end{figure*}

\subsection{Overview of Model Framework}

As illustrated in Fig.~\ref{framework}, the CurbNet framework comprises four main components: point cloud input with voxelization, feature extraction, feature aggregation to segmentation, and post-processing refinement.

In most road scenarios, curbs are primarily located at the junction between the road and the sidewalk. To enhance the model's ability to learn curb features, we introduced additional labels for the road and sidewalk during the training process. To align the input point cloud data with physical space characteristics (where the density of circularly scanned point clouds decreases with increasing distance), we partition the voxel blocks based on the point cloud distance. This approach minimizes the impact of uneven point cloud distribution.

During the feature extraction stage, voxelized features with dimensions \( F \times H \times W \times D \) are fed into a 5-layer deep encoder-decoder structure. Each encoder and decoder backbone incorporates the MSCA module, specifically designed for curb point cloud feature extraction. A stride-2 sparse convolution is employed as the pooling function.

Subsequently, the high-dimensional voxelized features (\( F \times H \times W \times D \)) are aggregated and then input into the segmentation head to obtain the detection results. Finally, the curb detection results undergo post-processing involving multi-clustering, fitting, and noise removal to further enhance detection accuracy.

The integration of these components within the CurbNet framework enables robust and precise curb detection, addressing the challenges posed by varying point cloud densities and the complex spatial characteristics of road environments.

\vspace{-5pt}
\subsection{Multi-Scale and Channel Attention (MSCA) Module}

\begin{figure}[t]
    \centering
    \includegraphics[width=0.5\textwidth]{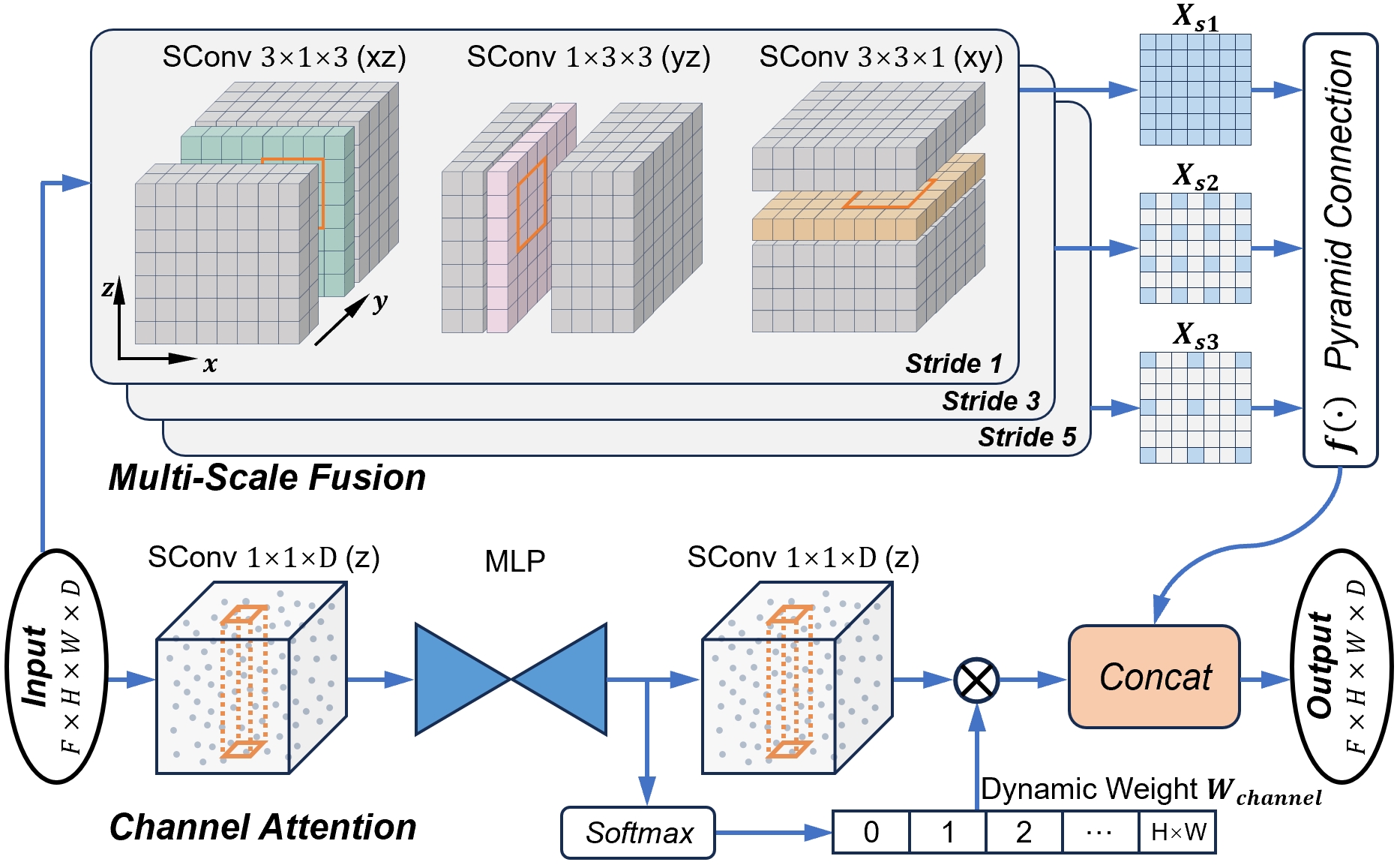}
    \vspace{-10pt}
    \caption{\textbf{Structure of multi-scale and channel attention (MSCA) module.} SConv means Sparse convolution layer. Multi-Scale Fusion is mainly used to fuse spatial features of different scales, and Channel Attention is used to dynamically extract height features of the z-axis.}
    \label{MSCA}
    \vspace{-15pt}
\end{figure}

The structure of the MSCA module, illustrated in Fig.~\ref{MSCA}, serves as a fundamental component of the encoder-decoder architecture. The MSCA module is composed of two main parts: multi-scale fusion and channel attention.

{\bf Multi-Scale Fusion.} Curbs in point clouds typically form a long curve along both sides of the road. However, as the LiDAR scanning distance increases, the point cloud density decreases, leading to sparser curb features. This results in significant scale differences in the features represented by the same number of point clouds at different distances after voxelization. To address this, we designed the MSCA module with a multi-scale fusion strategy. We employ convolution groups with different strides (1, 3, 5) to construct a feature pyramid, and then use dense pyramid connections to deeply fuse the output features of different scales. Inspired by text detection methods~\cite{wang2019shape}, we adopt three asymmetric sparse convolution kernels to target the regions in the xy, xz, and yz planes within the voxel space. This approach captures multi-dimensional curb feature information. Compared to traditional 3×3×3 convolutions, the combination of asymmetric sparse convolutions offers higher computational efficiency while maintaining the same receptive field.

Formally, given the input feature map $\mathbf{X} \in \mathbb{R}^{F \times H \times W \times D}$, where $F \times H \times W \times D$ denotes the spatial dimensions, we apply convolutions with varying strides to capture multi-scale features. Specifically, the input is processed using sparse convolutions (denoted as $\text{SConv}$) with stride values $s \in \{1, 3, 5\}$:
\begin{equation}
\begin{split}
    \mathbf{X}_{s1} &= \text{SConv}_{s=1}(\mathbf{X}), \\
    \mathbf{X}_{s3} &= \text{SConv}_{s=3}(\mathbf{X}), \\
    \mathbf{X}_{s5} &= \text{SConv}_{s=5}(\mathbf{X})
\end{split}
\end{equation}

These convolutions produce feature maps $\mathbf{X}_{s1}$, $\mathbf{X}_{s3}$, and $\mathbf{X}_{s5}$ at different spatial scales. 
The multi-scale outputs are fused as follows:
\begin{equation}
    \mathbf{X}_{\text{ms}} = f(\mathbf{X}_{s1}, \mathbf{X}_{s3}, \mathbf{X}_{s5})
\end{equation}

where $f(\cdot)$ represents the pyramid fusion operation, typically achieved by concatenating the outputs followed by additional sparse convolutions for deep integration.

{\bf Channel Attention.} The primary distinguishing feature of curbs is the subtle height difference between the road and the sidewalk, reflected in the z-axis of point cloud data. Unlike general point cloud segmentation algorithms such as Cylinder3D~\cite{zhou2020cylinder3d}, which primarily focus on feature learning in the xy plane, we designed the channel attention module to capture high-frequency features along the z-axis. Initially, a 1×1×D sparse convolution is applied to preliminarily extract channel features:
\begin{equation}
    \mathbf{C} = \text{SConv}_{1\times1\times D}(\mathbf{X})
\end{equation}

The channel features are then processed through an encoder-decoder structured MLP to further refine the height feature:
\begin{equation}
    \mathbf{C}_{\text{encoded}} = \text{MLP}_{\text{enc}}(\mathbf{C}), \quad \mathbf{C}_{\text{decoded}} = \text{MLP}_{\text{dec}}(\mathbf{C}_{\text{encoded}})
\end{equation}

The output of the MLP, $\mathbf{C}_{\text{decoded}}$, is then passed through two parallel branches. In the first branch, a softmax function generates dynamic weights for each channel:
\begin{equation}
    \mathbf{W}_{\text{channel}} = \text{softmax}(\mathbf{C}_{\text{decoded}})
\end{equation}

In the second branch, $\mathbf{C}_{\text{decoded}}$ is further processed by another 1×1×D sparse convolution:
\begin{equation}
    \mathbf{C}_{\text{conv}} = \text{SConv}_{1\times1\times D}(\mathbf{C}_{\text{decoded}})
\end{equation}

The $\mathbf{W}_{\text{channel}}$ are element-wise multiplied with the output $\mathbf{C}_{\text{conv}}$ to produce the final channel-attended feature map:
\begin{equation}
    \mathbf{X}_{\text{channel}} = \mathbf{W}_{\text{channel}} \odot \mathbf{C}_{\text{conv}}
\end{equation}

The final output of the MSCA is obtained by combining the multi-scale fusion output and the channel attention output:
\begin{equation}
    \mathbf{X}_{\text{output}} = \mathbf{X}_{\text{ms}} + \mathbf{X}_{\text{channel}}
\end{equation}

By combining multi-scale feature extraction with dynamic channel attention, the MSCA module effectively captures both spatial and height variations in point cloud data, thereby enhancing the model's ability to detect curbs.

\subsection{Loss Group}

In real-world scenarios, the point cloud data for curbs comprises only a small fraction compared to other categories such as roads and buildings. Using uniform loss weights can lead to training imbalances, adversely affecting the recognition performance for the minority class, i.e., curbs. To address this, we propose a novel combination of Adaptive Cross-Entropy (ACE) Loss and Lovász-Softmax Loss.

\noindent{\textit{{\textbf{1) Adaptive Cross Entropy (ACE) Loss}}}}

Due to the imbalance between the number of curb point clouds and the number of other categories such as roads and buildings. Standard loss functions like Cross Entropy (CE) \cite{zhang2018generalized} do not adequately address this imbalance, leading to suboptimal performance in recognizing minority classes. The standard CE loss is defined as:
\begin{equation}
\mathcal{L}_{CE}\left(p_{\mathrm{t}}\right) = -\log\left(p_{\mathrm{t}}\right)
\end{equation}

where \( p_t \) represents the predicted probability of the true class.

Given the disproportionate representation of classes in the point cloud data, we draw inspiration from the Focal Loss \cite{lin2017focal} to reallocate the loss contribution of easy and hard samples, significantly reducing the influence of the majority background samples:
\begin{equation}
\mathcal{L}_{FL}\left(p_{\mathrm{t}}\right) = -\alpha_{\mathrm{t}} \left(1 - p_{\mathrm{t}}\right)^\gamma \log\left(p_{\mathrm{t}}\right)
\end{equation}
The modulation factor \( (1 - p_t)^\gamma \) in Focal Loss is crucial as it down-weights the loss for well-classified examples and focuses learning on hard examples. However, Focal Loss treats all classes equally with the same modulation factor, which does not address the imbalance among foreground classes.

{\bf Adaptive Class-Wise Focusing Factor.} To tackle both the foreground-background imbalance and the inter-foreground class imbalance, we introduce an adaptive, class-wise focusing factor \( \gamma^i \) that adjusts according to the imbalance degree of each class \( i \). The adaptive focusing factor \( \gamma^i \) is defined as:
\begin{equation}
\begin{aligned}
\gamma^i & = \gamma_a + \gamma_b^i \\
& = \gamma_a + s\left(1 - \eta^i\right)
\end{aligned}
\end{equation}
Here, \( \gamma^i \) is decomposed into a class-agnostic parameter \( \gamma_a \) and a class-specific parameter \( \gamma_b^i \). The parameter \( \gamma_a \) represents the basic focusing factor under balanced data scenarios, while \( \gamma_b^i \geq 0 \) is a variable parameter related to the imbalance degree of class \( i \). 
The term \( \eta^i = N_i/N \) where \( N \) is the total number of points in the point cloud, and \( N_i \) is the number of points in class \( i \). 
The value of \( \eta^i \) is constrained to the range \([0, 1]\), and \( 1 - \eta^i \) inversely reflects the weight for low-frequency classes. The hyperparameter \( s \) is a scaling factor that determines the upper limit of \( \gamma^i \).

{\bf Dynamic Weight Factor.} While the adaptive focusing factor \( \gamma^i \) ensures more loss contribution from rare samples, it does not fully resolve the class imbalance problem. Therefore, we introduce a dynamic weighting factor \( \omega^i \) to provide higher weights for rare classes:
\begin{equation}
\omega^i = \frac{1}{\log(\delta + \eta^i)}
\end{equation}
where \( \delta \) is a small constant to prevent division by zero.

Combining these components, the final ACE Loss is expressed as:
\begin{equation}
\begin{aligned}
\mathcal{L}_{ACE}\left(p_{\mathrm{t}}\right) & = -\alpha_{\mathrm{t}} \omega^i \left(1 - p_{\mathrm{t}}\right)^{\gamma^i} \log\left(p_{\mathrm{t}}\right) \\
& = -\sum_{i=1}^C \alpha_{\mathrm{t}} \frac{1}{\log(\delta + \eta^i)} \left(1 - p_{\mathrm{t}}\right)^{\gamma_a + \gamma_b^i} \log\left(p_{\mathrm{t}}\right)
\end{aligned}
\end{equation}

The ACE Loss effectively prioritizes the learning of rare class samples by dynamically adjusting both the focusing factor and the class weights based on the distribution of point cloud data, thereby addressing the critical issue of class imbalance in curb detection.

\noindent{\textit{{\textbf{2) Lovász-Softmax Loss}}}}

Lovász Loss is particularly effective in handling imbalanced datasets and excels in addressing sparse boundary issues \cite{jadon2020survey}. Compared to traditional cross-entropy loss, it demonstrates superior performance in terms of Intersection over Union (IoU) scores. For a given true label vector \( \boldsymbol{y}^* \) and a predicted label vector \( \widetilde{\boldsymbol{y}} \), the IoU index for class c is defined as:

\begin{equation}
\text{IoU}_c\left(\boldsymbol{y}^*, \widetilde{\boldsymbol{y}}\right) = \frac{\left|\left\{\boldsymbol{y}^* = c\right\} \cap \left\{\widetilde{\boldsymbol{y}} = c\right\}\right|}{\left|\left\{\boldsymbol{y}^* = c\right\} \cup \left\{\widetilde{\boldsymbol{y}} = c\right\}\right|}
\end{equation}
This index provides the ratio between the intersection and union of the true and predicted masks within the range [0, 1], with the convention 0/0 = 1. The corresponding loss function employed in empirical risk minimization is:
\begin{equation}
\Delta_{\text{IoU}_c}\left(\boldsymbol{y}^*, \widetilde{\boldsymbol{y}}\right) = 1 - \text{IoU}_c\left(\boldsymbol{y}^*, \widetilde{\boldsymbol{y}}\right)
\end{equation}
For multi-label datasets, it is customary to average across classes, yielding the Mean IoU (mIoU).

The Lovász-Softmax loss extends this concept by applying the Lovász extension to the softmax probabilities of a model's output. It optimizes a convex surrogate of the IoU score, which is more suitable for gradient-based optimization. Specifically, the loss $\mathcal{L}_{IoU}$ for a set of classes \( C \) is defined as:
\begin{equation}
\mathcal{L}_{IoU}(\boldsymbol{y}^*, \widetilde{\boldsymbol{y}}) = \sum_{c \in C} \Delta_{\text{IoU}_c}\left(\boldsymbol{y}^*, \widetilde{\boldsymbol{y}}\right)
\end{equation}
The computation involves ordering the pixels by error margin and computing a weighted sum of the individual errors, thus directly targeting the errors that most impact the IoU score.


\subsection{Multi-Cluster and Curve Fitting}

\begin{figure}[t]
    \centering
    \includegraphics[width=0.45\textwidth]{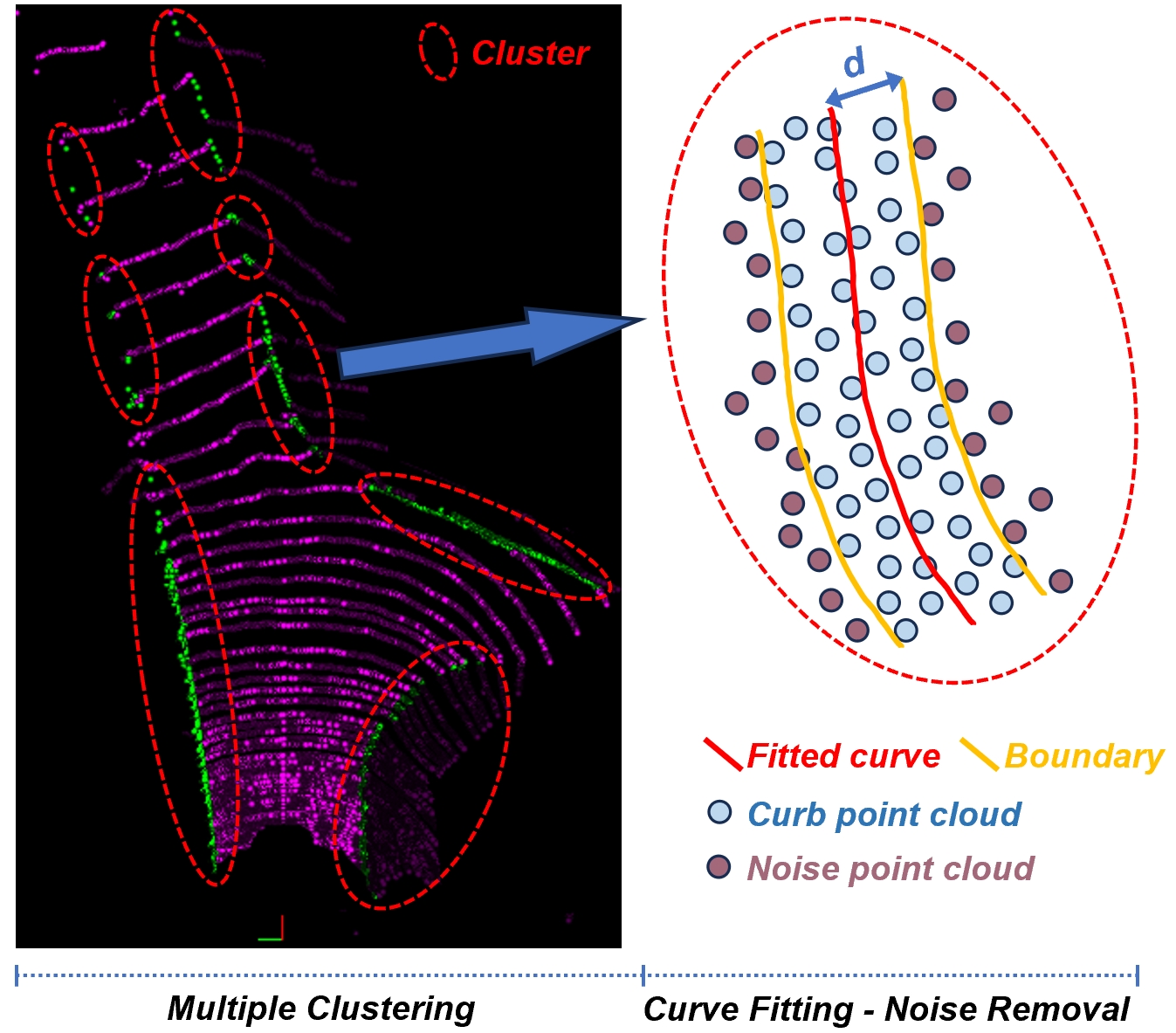}
    \vspace{-5pt}
    \caption{\textbf{Process of multiple clustering and fitting to remove noise points.}The left figure shows the effect of multiple clustering in discontinuous scenes. The right figure shows the method of curve fitting and setting distance to remove noise points.}
    \label{cluster}
    \vspace{-15pt}
\end{figure}

This paper introduces a post-processing method based on multi-cluster refitting to filter noise points from LiDAR data segmentation results, thereby enhancing the detection accuracy of curbs. Due to the increasing sparsity of LiDAR point clouds with distance and the potential interruption of curb lines due to obstructions, direct curb clustering along the sides of roads is challenging, as shown in Fig.~\ref{cluster}. Thus, we adopt a multi-cluster strategy, treating the curb in multiple segments.

To address this challenge, we initially apply the Density-Based Spatial Clustering of Applications with Noise (DBSCAN) algorithm \cite{schubert2017dbscan} for preliminary segmentation of the detected curbs. DBSCAN is characterized by its ability to identify clusters of arbitrary shapes without a predefined number of clusters, efficiently handling noise points. The core idea of DBSCAN revolves around setting a neighborhood radius $\varepsilon$ (Eps) and a minimum sample number minPts (min-samples) to determine cluster membership. In our study, we set Eps to 1 and min-samples to 5.

Let $P$ be a point in the point cloud; its $\varepsilon$-neighborhood, denoted as $N_{\varepsilon}(P)$, is defined as:

\begin{equation}
N_{\varepsilon}(P)=\{Q \in \text { Dataset } \mid \operatorname{dist}(P, Q) \leq \varepsilon\}
\end{equation}

where $\operatorname{dist}(P, Q)$ represents the distance between points $P$ and $Q$. $P$ is considered a core point if its $\varepsilon$-neighborhood contains at least minPts points, i.e., $\left|N_{\varepsilon}(P)\right| \geq \operatorname{minPts}$.

Post-clustering, we fit polynomial curves to each independent curb segment. The key is to precisely fit the geometric shape of the curb while eliminating noise points not belonging to the curb. After polynomial curve fitting of each segment, by calculating the distance from points to the fitted curve, we can effectively identify and eliminate noise points located outside the fitted curve, as shown in Fig.~\ref{cluster}.

During the polynomial curve fitting, the aim is to eliminate noise points not belonging to the curb. Assuming the curve equation is $f(x)$, for any point $P(x, y)$ in the point cloud, we calculate the perpendicular distance $d$ to the curve:
\begin{equation}
d(P, f)=|y-f(x)|
\end{equation}

If $d(P, f)$ exceeds a predetermined threshold $\delta$, the point P is considered noise and is removed from the dataset:
\begin{equation}
\text { If } d(P, f)>\delta, \text { then } \mathrm{P} \text { is noise }
\end{equation}

The complete operation process is shown in Algorithm.~\ref{algorithm1}. In order to improve calculation efficiency, we use parallel computing and KDTree's efficient point cloud search scheme. Through this approach, combining the DBSCAN algorithm with polynomial curve fitting effectively identifies and extracts accurate curb lines, while eliminating noise points, thus improving the overall detection accuracy.

\begin{algorithm}
\caption{Multi-Cluster and Curve Fitting Post-Processing}
\label{algorithm1}
\begin{algorithmic}[1]
\Require Point cloud data, $\varepsilon$, minPts, $\delta$
\Ensure Refined curb line segmentation
\State \textbf{Step 1:} Apply DBSCAN to point cloud data with parallel processing
\For{each point $P$ in point cloud}
    \State Compute $N_{\varepsilon}(P)$ in parallel computation
    \If{$\left|N_{\varepsilon}(P)\right| \geq \text{minPts}$}
        \State Mark $P$ as a core point
    \EndIf
\EndFor
\State \textbf{Step 2:} Segment curb lines into clusters
\State \textbf{Step 3:} Fit polynomial curves in batch processing
\For{each curb segment}
    \State Fit polynomial curve $f(x)$
    \For{each point $P(x, y)$ in segment}
        \State Calculate perpendicular distance $d(P, f)$
        \State Build \textbf{KDTree} for efficient nearest neighbor search
        \If{$d(P, f) > \delta$}
            \State Remove $P$ as noise
        \EndIf
    \EndFor
\EndFor
\State \textbf{Step 4:} Output refined curb line segments
\end{algorithmic}
\end{algorithm}
\vspace{-5pt}

\section{Experiment and Analysis}

\subsection{Experiment Setup}
\noindent{\textit{{\textbf{1) Training Details}}}}

Our model training was conducted in an Ubuntu 20.04 environment, utilizing an Intel(R) Xeon(R) Gold 5318S CPU @ 2.10GHz and an NVIDIA RTX 3090 GPU. We employed the PyTorch framework for model training and set training parameters with a batch size of 6, a total of 100 epochs, and a learning rate of 0.001.

Regarding the datasets, we employed two distinct datasets for training and evaluation: the publicly available NRS dataset and our custom-built 3D-Curb dataset. Both datasets comprised curb data collected from forward-facing trajectories, each extending over 40 meters. To enhance feature learning for curb detection, we augmented the training process by including two additional categories: road and sidewalk.

\noindent{\textit{{\textbf{2) Evaluation Metrics}}}}

The performance of our curb detection method was rigorously evaluated using standard metrics. These include Precision, Recall, and F-1 score, which are quintessential for quantifying the accuracy and reliability of classification models. Precision, defined as $\frac{TP}{TP + FP}$, measures the proportion of correctly predicted positive observations to the total predicted positives. Recall, calculated as $\frac{TP}{TP + FN}$, assesses the proportion of actual positives that were correctly identified. The F-1 score, given by $2 \times \frac{\text{Precision} \times \text{Recall}}{\text{Precision} + \text{Recall}}$, harmonizes the balance between Precision and Recall, providing a single measure of efficacy. Here, TP (True Positives) represents the number of correct positive predictions, FP (False Positives) denotes the count of negative instances incorrectly classified as positive, TN (True Negatives) refers to the count of correct negative predictions, and FN (False Negatives) signifies the instances where positive cases were wrongly predicted as negative. These metrics collectively offer a comprehensive view of our model's performance, which is crucial for the validation.

\subsection{Quantitative Results of Curb Detection}

\noindent{\textit{{\textbf{1) Model Training Results}}}}

In the NRS dataset experiments (refer to Table~\ref{Table2}), this study compared classic segmentation algorithms such as PointPillars, U-Net, Swin-Transformer, and CSWin-Transformer, as well as the state-of-the-art curb detection model, LCDeT. Leveraging the specially designed MSCA module for 3D curb scenarios, CurbNet achieved the highest detection performance on the NRS dataset. With the auxiliary training of relevant labels, CurbNet attained Precision, Recall, and F-1 scores of 0.8281, 0.8329, and 0.8308, respectively.
Among them, auxiliary training helps improve Precision by 0.5 points, and post-processing helps improve Precision by 1.4 points.

\begin{table}[t!]
\renewcommand\arraystretch{1.5}
\caption{COMPARISON OF RESULTS IN NRS-DATASET \cite{gao2023lcdet}. (w/o) and (w/) represent WITHOUT or WITH the auxiliary training of road and sidewalk labels respectively.}
\centering
\footnotesize
\begin{tabular}{ccccc}
\hline Method & Precision & Recall & F-1 score \\
\hline
PointPillars\cite{lang2019pointpillars} & 0.759 & 0.6019 & 0.6524 \\
U-Net\cite{ronneberger2015u} & 0.7546 & 0.7018 & 0.7172 \\
Swin-T\cite{liu2021swin} & 0.7216 & 0.7034 & 0.7001 \\
CSWin-T\cite{dong2022cswin} & 0.7692 & 0.6597 & 0.6966 \\
LCDeT\cite{gao2023lcdet} & 0.8257 & 0.8050 & 0.8092 \\
CurbNet (w/o) & 0.8225 & 0.8264 & 0.8234 \\
CurbNet (w) & \underline{0.8281} & \underline{0.8329} & \underline{0.8308} \\
CurbNet-post & \textbf{0.8420} & \textbf{0.8496} & \textbf{0.8457} \\
\hline
\end{tabular}
\label{Table2}
\vspace{-5pt}
\end{table}

The experiments in the 3D-Curb dataset not only compared classic and advanced deep learning model algorithms but also included three traditional methods of manual feature extraction, as shown in Table~\ref{Table3}. Owing to the robust automatic feature extraction capabilities, deep learning methods significantly outperformed in curb detection, surpassing the other methods by over 10 points in Precision, Recall, and F-1 score. Among the deep learning methods, CurbNet surpassed the best-performing supervised learning model on the SemanticKITTI dataset, Cylinder3D, by 2.5 points in Precision, and exceeded the knowledge distillation model PVKD by 1.5 points. 
Among them, Precision is improved by more than 1 point through auxiliary training, and by 4.5 points through post-processing.


\begin{table}[t!]
\renewcommand\arraystretch{1.5}
\caption{COMPARISON OF RESULTS IN 3D-Curb DATASET. (w/o) and (w/) represent WITHOUT or WITH the auxiliary training of road and sidewalk labels respectively.}
\centering
\footnotesize
\begin{tabular}{>{\centering\arraybackslash}m{1.5cm}cccc}
\hline Difference & Methods & Precision & Recall & F-1 score \\
\hline \multirow{3}{*}{Manual Feature} & Zhang\cite{zhang2018road} & 0.6854 & 0.6564 & 0.6053 \\
& Sun \cite{sun20193d} & 0.6878 & 0.6864 & 0.6297 \\
& Wang\cite{wang2020speed} & 0.7209 & 0.7013 & 0.6973\\
\hline
\multirow{6}{*}{Deep Learning} & 3D U-Net\cite{cciccek20163d} & 0.7695 & 0.7492 & 0.7592 \\
& Cylinder3D \cite{zhou2020cylinder3d} & 0.8049 & 0.8038 & 0.7942 \\
& PVKD \cite{hou2022point} & 0.8125 & 0.8089 & 0.8025 \\
& CurbNet (w/o) & 0.8178 & 0.8395 & 0.8325 \\
& CurbNet (w/) & \underline{0.8292} & \underline{0.8567} & \underline{0.8427} \\
& CurbNet-post & \textbf{0.8743} &  \textbf{0.8647} & \textbf{0.8695} \\
\hline
\end{tabular}
\label{Table3}
\vspace{-10pt}
\end{table}

\noindent{\textit{{\textbf{2) Tolerance Results}}}}

Since curbs resemble elongated curves, relevant research often further tests model performance within a certain error range. Experiments are typically conducted in meters and pixels, with 1 pixel approximately equal to 0.1m of Tolerance, and common experimental settings range from 0.1m to 0.4m.

In our study, we conducted Tolerance performance tests on the 3D-Curb dataset, setting four Tolerances ranging from just 0.05m to 0.2m, as shown in Table~\ref{tolerabce}. Tests were carried out on four models: 3D U-Net, Cylinder3D, PVKD, and CurbNet (ours). With the increase in error Tolerance, performance metrics improved significantly. At 0.05m Tolerance, Precision improved by an average of 4 points; at 0.1m, by 9 points; at 0.15m, by 12 points; and at 0.2m, by 13 points. As Tolerance increased, performance gains gradually reached saturation. Notably, our CurbNet exceeded 0.95 in the average values of Precision, Recall, and F-1 score at just 0.15m Tolerance. This represents the optimal performance achieved in curb detection based on point cloud segmentation.

\begin{table*}[t]
\renewcommand\arraystretch{1.5}
\caption{Comparison of different tolerances in 3D-Curb Dataset. (w/o) and (w/) represent WITHOUT or WITH the auxiliary training of road and sidewalk labels respectively.}
\vspace{-5pt}
\centering
\begin{tabular}{>{\centering\arraybackslash}m{2.3cm}cccccccccccccc}
\hline  
  & \multicolumn{4}{c}{ Precision } & & \multicolumn{4}{c}{ Recall } & & \multicolumn{4}{c}{ F-1 score } \\
\cline{2-5} \cline{7-10} \cline{12-15} 
Tolerance (m) & 0.05 & 0.10 & 0.15 & 0.20 & & 0.05 & 0.10 & 0.15 & 0.20 & & 0.05 & 0.10 & 0.15 & 0.20 \\
\hline 
3D U-Net\cite{cciccek20163d} & 0.8293 & 0.8583 & 0.8881 & 0.9012 & & 0.8078 & 0.8498 & 0.8745 & 0.8843 & & 0.7933 & 0.8389 & 0.8662 & 0.8777 \\
Cylinder3D\cite{zhou2020cylinder3d} & 0.8403 & 0.8909 & 0.9221 & 0.9360 & & 0.8588 & 0.9007 & 0.9253 & 0.9355 & & 0.8391 & 0.8856 & 0.9136 & 0.9257 \\
PVKD \cite{hou2022point} & 0.8422 & 0.8898 & 0.9264 & 0.9398 & & 0.8595 & 0.9021 & 0.9305 & 0.9417 & & 0.8419 & 0.8902 & 0.9188 & 0.9343 \\
CurbNet (w/o) & 0.8588 & 0.9091 & 0.9397 & 0.9503 & & 0.8934 & \underline{0.9355} & \underline{0.9559} & \underline{0.9689} & & 0.8793 & 0.9211 & 0.9476 & 0.9597 \\
CurbNet (w/) & \underline{0.8667} & \underline{0.9158} & \underline{0.9461} & \underline{0.9587} & & \textbf{0.9029} & \textbf{0.9434} & \textbf{0.9663} & \textbf{0.9760} & & \underline{0.8845} & \underline{0.9295} & \textbf{0.9562} & \textbf{0.9673} \\
CurbNet-post & \textbf{0.9139} & \textbf{0.9530} & \textbf{0.9679} & \textbf{0.9738} & & \underline{0.8976} & 0.9303 & 0.9441 & 0.9501 & & \textbf{0.9056} & \textbf{0.9413} & \underline{0.9558} & \underline{0.9617} \\
\hline
\end{tabular}
\label{tolerabce}
\end{table*}

\subsection{Visualization Results of Curb Detection}

\begin{figure*}[t!]
    \centering
    \includegraphics[width=.95\textwidth]{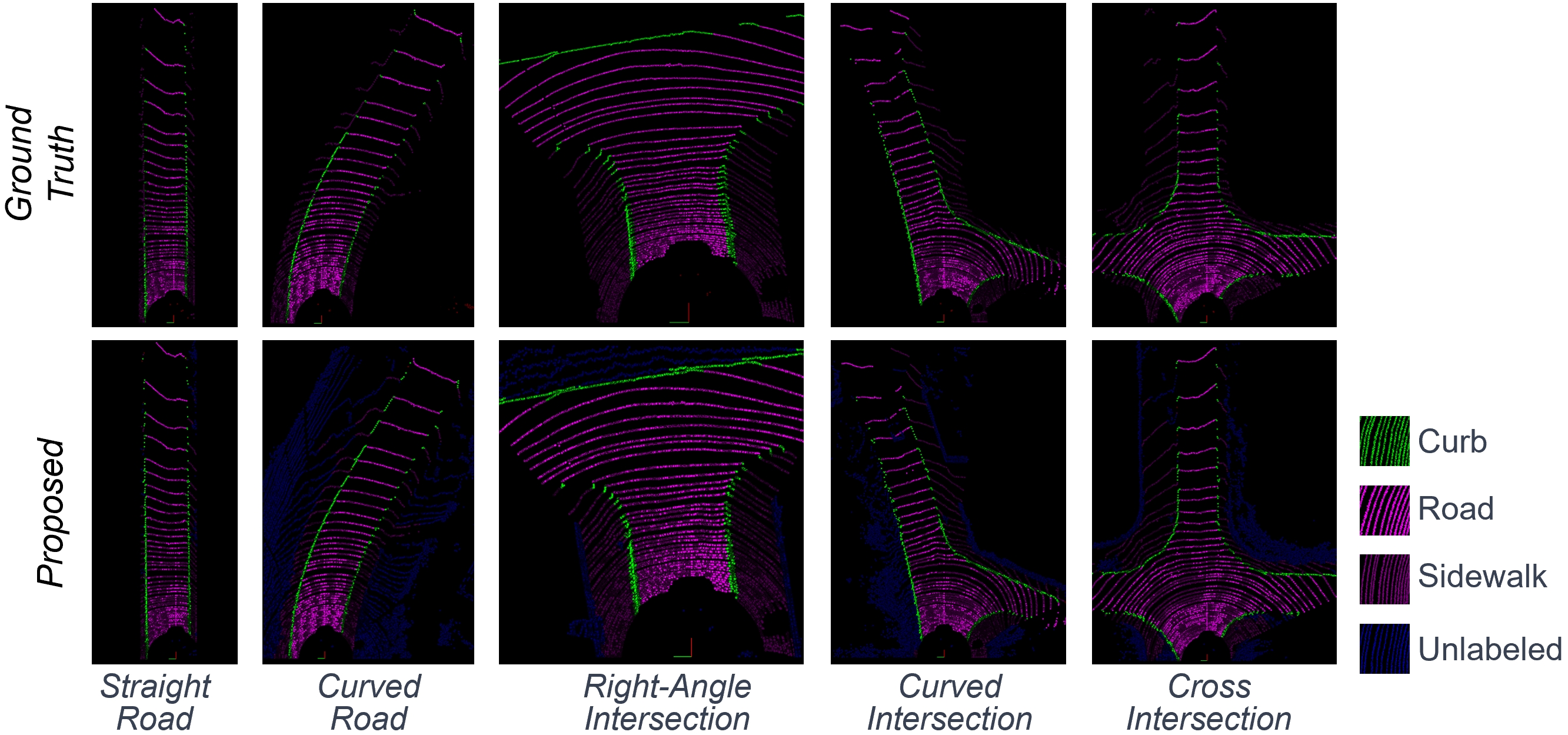}
    \vspace{-5pt}
    \caption{\textbf{Curb detection result in 3D-Curb dataset.} We compared the curb detection results at five classic intersections. The model accurately detected the curb area and was even better than the ground truth on the curve road.}
    \label{3d-curb-result}
    \vspace{-5pt}
\end{figure*}

We conducted a visual analysis of the test results obtained using our CurbNet model, comparing them with the ground truth to further validate the model's detection performance. Fig.~\ref{3d-curb-result} illustrates the visualization results in scenarios without obstructions, showcasing the curb detection results in five common road scenes: straight road, curved road, right-angle intersection, curved intersection, and cross intersection. The images clearly demonstrate that the CurbNet model successfully identified all curb lines present in the ground truth, even in complex intersection scenarios. Notably, in the curved intersection scenario, the results identified by CurbNet were even more precise than the ground truth annotations, showcasing our model's exceptional feature extraction capabilities and scene generalizability.

\begin{figure*}[t!]
    \centering
    \includegraphics[width=.95\textwidth]{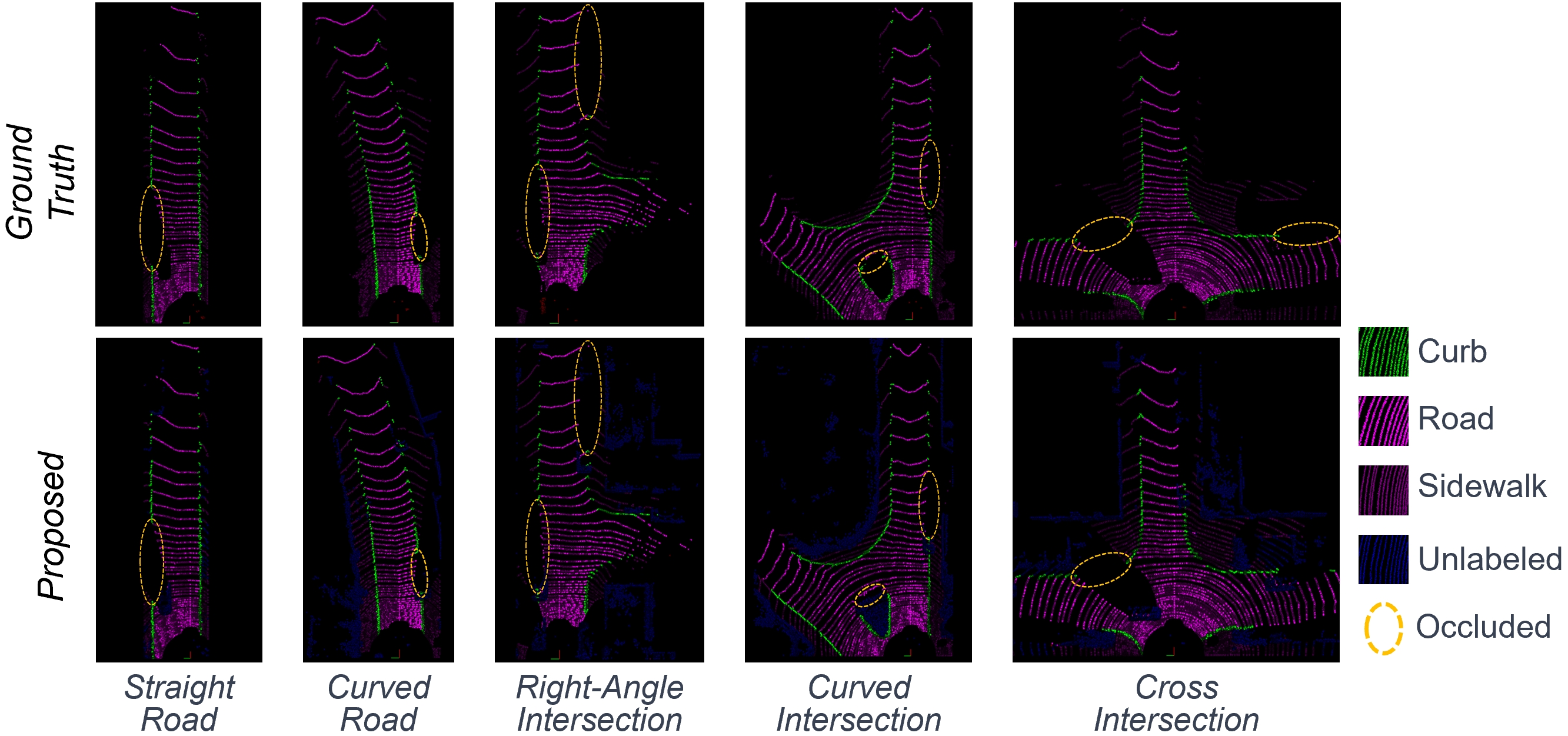}
    \vspace{-5pt}
    \caption{\textbf{Curb detection result in 3D-Curb dataset under occlusion.} We compared the curb detection results at five classic intersections with occlusion. Even under occlusion, it does not affect curb detection in other areas.}
    \label{3d-curb-result-occ}
    \vspace{-5pt}
\end{figure*}

Fig.~\ref{3d-curb-result-occ} also displays the visual recognition results in scenarios with obstructions, where yellow dashed circles highlight the obstructed areas. Despite the absence of point clouds in obstructed areas, the model accurately identified curbs in the other parts without being influenced by the obstructed regions. As long as the input data contained curb features, CurbNet could accurately detect them, unaffected by the obscured areas.

\begin{figure*}[t!]
    \centering
    \includegraphics[width=.99\textwidth]{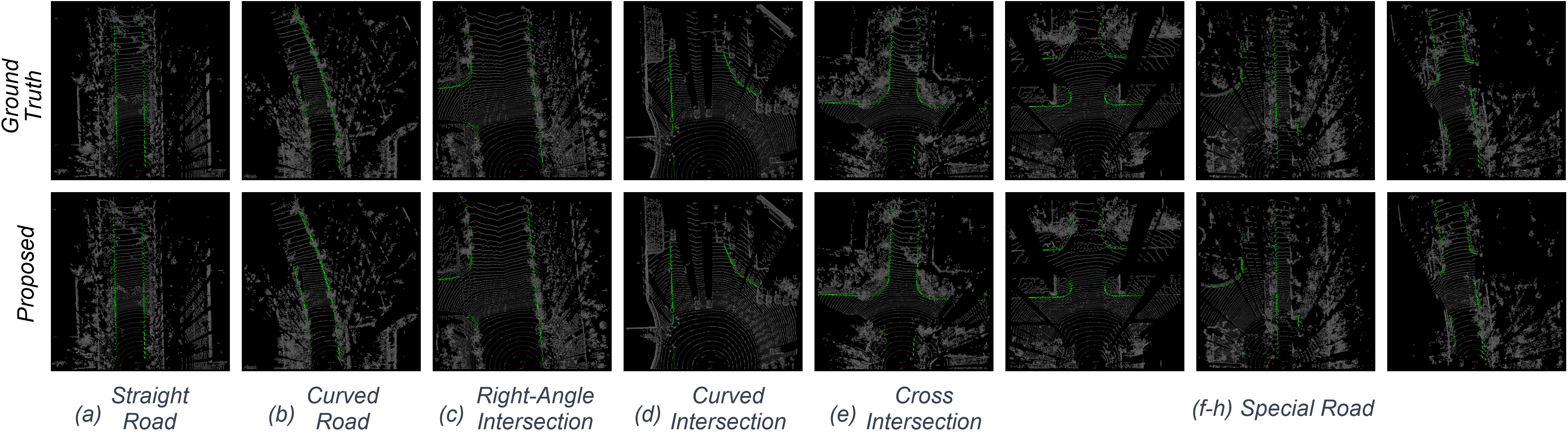}
    \vspace{-5pt}
    \caption{\textbf{Curb detection result in NRS Dataset.} The excellent detection performance and generalization of the proposed method are well demonstrated on the NRS dataset, and accurate detection can be performed even at the special intersection (f-h).}
    \label{nrs-result}
    \vspace{-6pt}
\end{figure*}

Furthermore, we conducted a visual analysis of the results on the NRS dataset, as shown in Fig.~\ref{nrs-result}. This analysis primarily showcases curb detection in five common road scenarios and three special intersection types. Through a comparative visualization of the detection results and ground truth, our method accurately identified the respective curb features.
The NRS dataset includes several challenging scenes characterized by irregular road structures and unique curb configurations, such as sharp elevation changes, occlusions, and narrow pathways. As shown in Fig.~\ref{nrs-result} (f)-(h), our method successfully detected the respective curb features in these biased scenarios. This demonstrates the robustness of CurbNet in handling dataset bias and its ability to generalize effectively to complex and irregular road conditions. The consistent detection results in these scenarios validate the model's capability to adapt to diverse real-world environments, highlighting its potential for broader applicability in intelligent driving systems.

\subsection{Post-Processing Experiment}

\begin{figure*}[t!]
    \centering
    \includegraphics[width=.99\textwidth]{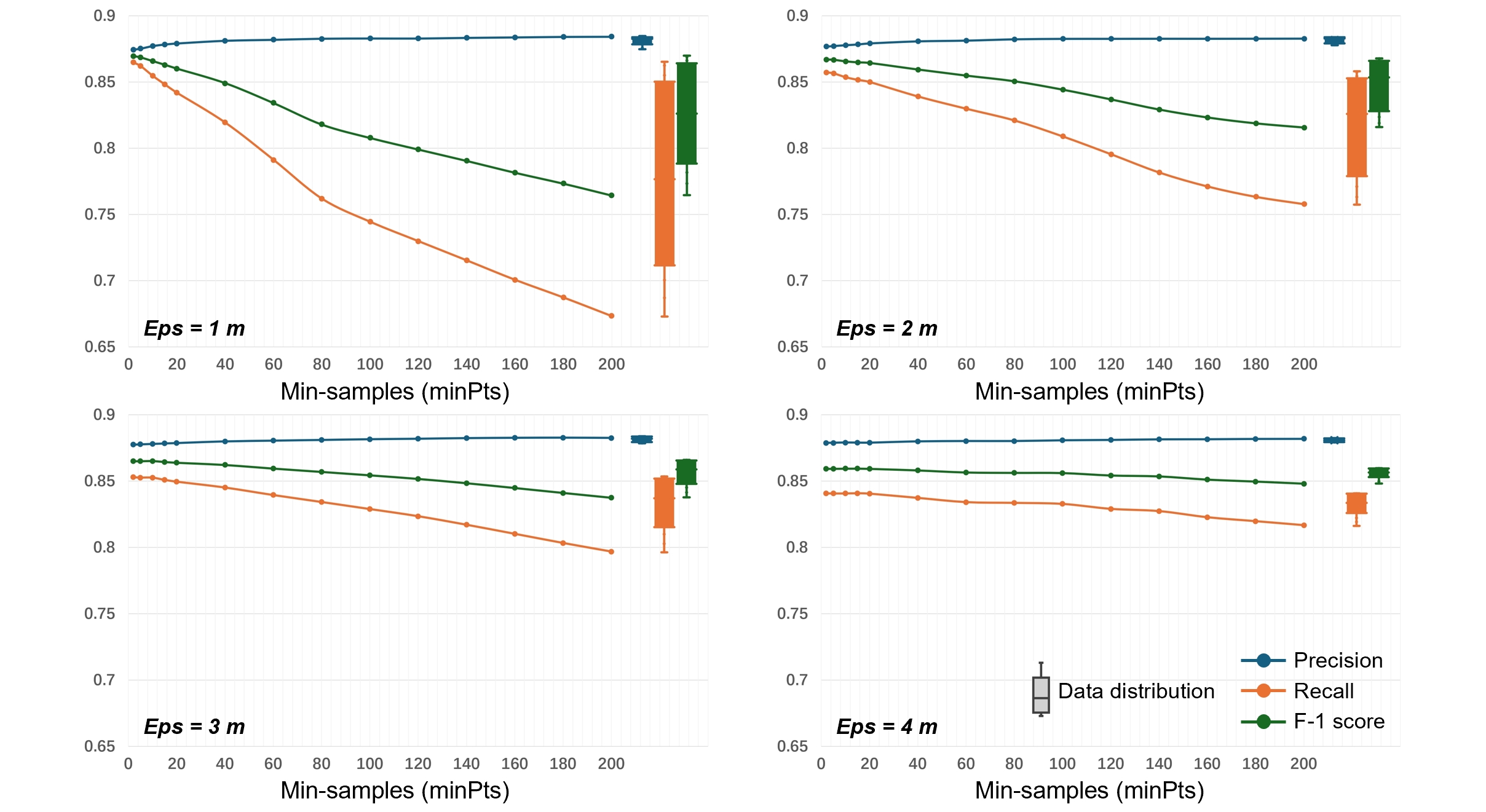}
    \vspace{-8pt}
    \caption{\textbf{Parameter adjustment experiments for multiple clustering and fitting.} We conduct control variable experiments on the main parameter distance variable Eps and the minimum sample point variable minPts of DBSCAN clustering.}
    \label{cluster-result}
    \vspace{-10pt}
\end{figure*}

In this paper, we conducted controlled experiments to compare the proposed post-processing method. Given the use of multiple clustering followed by curve fitting in post-processing, the setting of clustering parameters plays a crucial role in its effectiveness. Based on the road width and point cloud density characteristics of the 3D-Curb dataset, we experimented with varying the distance variable Eps (from 1m to 4m) and the minimum sample points variable minPts (from 2 to 200), as illustrated in Fig.~\ref{cluster-result}.

As the clustering distance Eps increases, changes in the performance metrics of post-processing become more gradual and similar. However, when the minimum sample points minPts are lower, the performance decreases compared to when Eps is 1m. This is attributed to the larger curb clustering caused by greater clustering distances, resulting in minimal changes post-curve fitting. Additionally, larger clusters tend to overlook sparsely distributed point clouds during curve fitting, thereby reducing performance.

Fig.~\ref{cluster-result} clearly demonstrates that as the minimum sample points variable minPts increases, the Precision metric of post-processing gradually improves, but both Recall and F-1 score metrics significantly decrease, especially at Eps settings of 1m and 2m. This decline is due to the increase in the number of minPts leading to the neglect of sparsely distributed curb point clouds at greater distances, resulting in a noticeable drop in Recall and F-1 scores.

Based on comparative experimental results and trade-off between metrics, smaller Eps distances and fewer minPts numbers yield the most optimal post-processing outcomes.

\subsection{Ablation Study}

As shown in Table~\ref{ablation}, this study conducted comparative ablation experiments focusing on the main loss functions and crucial module designs of our model. We first conducted individual experiments on the employed $\mathcal{L}_{CE}$ (Cross-Entropy Loss), $\mathcal{L}_{FL}$ (Focal Loss), $\mathcal{L}_{ACE}$ (Adaptive Cross-Entropy Loss), and $\mathcal{L}_{IoU}$ (Intersection over Union Loss). Subsequently, we tested combinations of $\mathcal{L}_{CE}$+$\mathcal{L}_{IoU}$ loss, $\mathcal{L}_{FL}$+$\mathcal{L}_{IoU}$ loss and $\mathcal{L}_{ACE}$+$\mathcal{L}_{IoU}$ loss. In the experimental results, due to the design of the $\mathcal{L}_{ACE}$ loss addressing the imbalance in the number of curb point clouds compared to other categories, its disproportionate weight settings caused the model to overly focus on recall during training. However, the interaction with the $\mathcal{L}_{IoU}$ loss led to a more balanced overall performance, achieving optimal detection capabilities. 
The combination of $\mathcal{L}_{ACE}$+$\mathcal{L}_{IoU}$ loss outperformed the $\mathcal{L}_{CE}$+$\mathcal{L}_{IoU}$ loss group by 1.3 points in Precision and 2.2 points in Recall. It also surpassed the $\mathcal{L}_{FL}$+$\mathcal{L}_{IoU}$ loss group with improvements of 0.7 points in Precision and 2 points in Recall.


Finally, we compared the model's performance with and without the MSCA module. When the MSCA module was not utilized, the model relied on the original encoder-decoder structure from Cylinder3D \cite{zhou2020cylinder3d}. Under the $\mathcal{L}_{ACE}$+$\mathcal{L}_{IoU}$ loss setting, the inclusion of the MSCA module significantly enhanced the performance metrics, particularly increasing Recall by 2 points and the F-1 score by 1 point, which further demonstrates the MSCA module's effectiveness in improving curb detection performance.

\begin{table}
\renewcommand\arraystretch{1.5}
\caption{Ablation of different loss functions and modules}
\vspace{-8pt}
\centering
\setlength{\tabcolsep}{2mm}
\footnotesize
\begin{tabular}{@{}cccccccc>{\columncolor{gray!20}}c@{}}
\hline
 $\mathcal{L}_{CE}$ & $\mathcal{L}_{FL}$ & $\mathcal{L}_{ACE}$ & $\mathcal{L}_{IoU}$ & MSCA & Precision & Recall & F-1 \\
\hline
\ding{51} & & & & \ding{51}& 0.8174 & 0.8303 & 0.8238 \\
 & \ding{51} & & & \ding{51}& 0.8196 & 0.8346 & 0.8273 \\
& & \ding{51} & & \ding{51}& 0.7933 & \textbf{0.8707} & 0.8355 \\
& & & \ding{51} & \ding{51}& 0.8234 & 0.8367 & 0.8339 \\
\ding{51} & &  & \ding{51} & \ding{51}& 0.8186 & 0.8472 & 0.8374 \\
 & \ding{51} &  & \ding{51} & \ding{51}& 0.8241 & 0.8498 & \underline{0.8401} \\
\hline
& & \ding{51}  & \ding{51} & & \underline{0.8297} & 0.8496 & 0.8395 \\
& & \ding{51}  & \ding{51} & \ding{51}& \textbf{0.8311} & \underline{0.8695} & \textbf{0.8499} \\
\hline
\end{tabular}
\label{ablation}
\end{table}

\subsection{Real Scene Experiment}
\begin{figure}[t]
    \centering
    \includegraphics[width=0.5\textwidth]{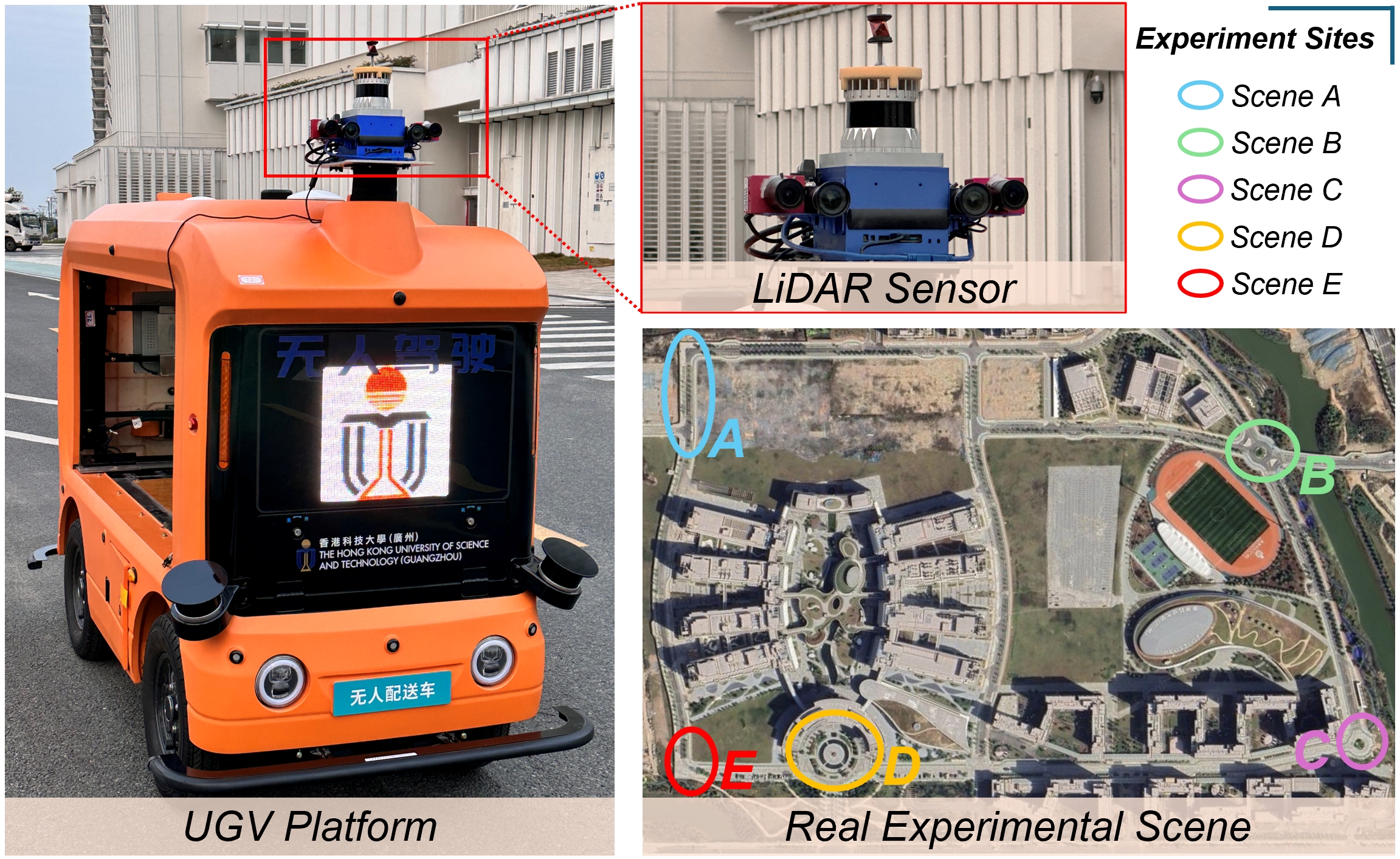}
    \vspace{-10pt}
    \caption{\textbf{Setup of real scene experiment.} We used the delivery vehicle equipped with LiDAR to conduct real scene experiments. There are a total of five experimental sections distributed on the HKUST Guangzhou campus.}
    \label{realscene-set}
    \vspace{-8pt}
\end{figure}

To further validate the performance and effectiveness of the proposed method, we conducted real-world experiments in addition to the dataset experiments. As depicted in Fig.~\ref{realscene-set}, we utilized an autonomous delivery vehicle as our experimental platform, equipping it with a LiDAR sensor system mounted on its top. The LiDAR used was the OS1-U model with 128 lines, manufactured by Ouster. To ensure the generalizability of our experiments, the autonomous vehicle was driven on roads within the HKUST Guangzhou campus. Data collection and curb detection were carried out in five different road segments (Scene A, B, C, D, and E), as shown in the map in Fig.~\ref{realscene-set}. These segments included both standard road scenarios (straight roads, bends, and intersections) and complex ones (roundabout turns and intricate junctions).

\begin{figure*}[ht!]
    \centering
    \includegraphics[width=.99\textwidth]{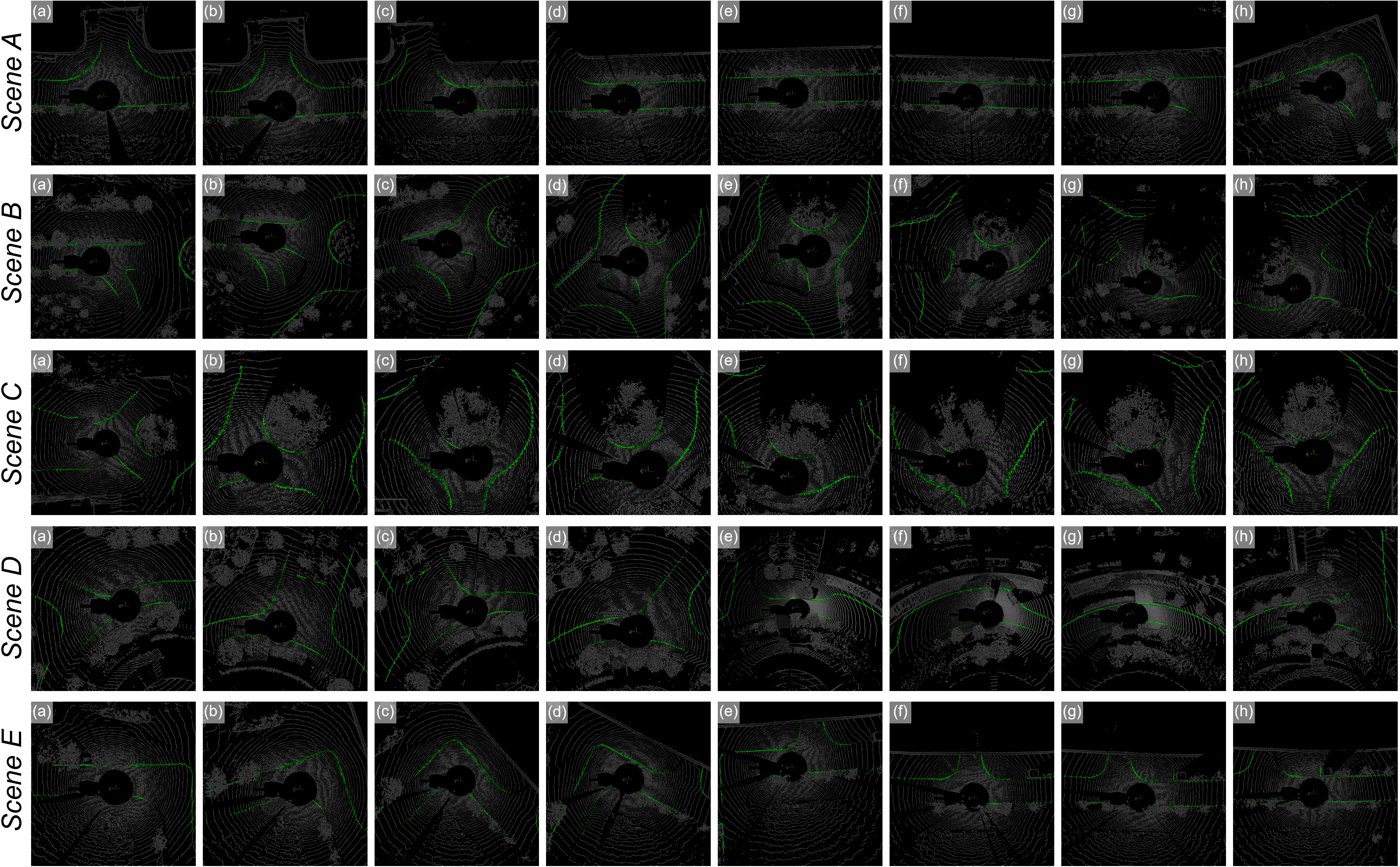}
    \vspace{-10pt}
    \caption{\textbf{Curb detection results of real experiments in five field scenarios.} Our method achieves excellent curb detection results in five real-world scenarios. In particular, it also shows excellent performance in complex intersection scenarios B and C with irregular curb distribution.}
    \label{real-exp}
    \vspace{-12pt}
\end{figure*}

The curb detection results in the 5 scenes are illustrated in Fig.~\ref{real-exp}, where we selected 8 representative images from each scene for visual analysis.
Overall, our method achieved commendable curb detection results in each scene, further demonstrating the robust curb feature extraction capability of the CurbNet model. In individual cases, our proposed method accurately detected not only the evident, extensive curbs, but also achieved remarkable results on small-scale curbs, which are typically less conspicuous and easily overlooked, as seen in Scene B (image d) and Scene D (images b and c). This success can be attributed to the model's design focusing on multi-scale feature fusion and height channel feature extraction. 
Similarly, our method also exhibited superior detection performance in complex intersection scenarios with irregular curb distributions (Scene B and Scene C). Particularly in Scene B, the method precisely detected all curbs present in the LiDAR point cloud.

Finally, we also quantitatively evaluate the real scene experiments by testing key metrics, as shown in Table~\ref{Table6}. By comparing with manually annotated ground truth, CurbNet achieved an average Precision, Recall, and F1 score of 0.8462, 0.8443, and 0.8453, respectively, across the five scenes. Among them, the Recall and F-1 score indicators obtained in Scene A are the highest, which are 0.8821 and 0.8569 respectively, and the Precision indicator obtained in Scene B is the highest 0.8950. These results corroborate with those obtained from dataset testing, further substantiating the excellent performance and generalizability of the CurbNet.

\begin{table}[t!]
\renewcommand\arraystretch{1.5}
\caption{RESULTS IN FIVE REAL SCENES.}
\centering
\footnotesize
\begin{tabular}{ccccc}
\hline Site & Precision & Recall & F-1 score \\
\hline
Scene A & 0.8331 & \textbf{0.8821} & \textbf{0.8569} \\
Scene B & \textbf{0.8950} & 0.8123 & 0.8517 \\
Scene C & \underline{0.8610} & 0.8366 & \underline{0.8530} \\
Scene D & 0.8018 & \underline{0.8533} & 0.8268 \\
Scene E & 0.8579 & 0.8388 & 0.8483 \\
\hline
All Scene & 0.8462 & 0.8443 & 0.8453 \\
\hline
\end{tabular}
\label{Table6}
\vspace{-6pt}
\end{table}

\subsection{Time Consumption}
In practical applications for intelligent vehicles, real-time curb detection is critical. As shown in Table~\ref{Table7}, we evaluated the time consumption for both the model inference and post-processing stages of our curb detection framework. To ensure the efficiency of the entire processing framework, the model inference is executed on the GPU, while the post-processing runs on the CPU.

In various scenarios, since the LiDAR used and the number of input point clouds are different, the time consumption of computational processing is also different. Nevertheless, the CurbNet framework consistently achieves an overall real-time performance exceeding 15 Hz during both model inference and post-processing stages. Notably, the post-processing stage operates faster than the model inference stage, ensuring the smooth and efficient operation of the entire framework.


Furthermore, we evaluated the real-time performance of CurbNet on an on-board processing unit, specifically the NVIDIA Jetson AGX Orin \cite{nvidia2022jetson}, which provides up to 275 TOPS of computational power with INT8 precision. Through model optimization using TensorRT and INT8 precision inference, the CurbNet achieved a processing speed exceeding 20 FPS on this device. These results underscore the feasibility of deploying CurbNet on modern edge computing platforms, ensuring real-time curb detection even under resource-constrained conditions. This highlights the practicality of CurbNet for real-world autonomous driving applications.

\begin{table}[t!]
\renewcommand\arraystretch{1.5}
\caption{Comparison of time consumption in model inference and post-processing, the unit is (FPS/ms).}
\centering
\footnotesize
\begin{tabular}{ccccc}
\hline Stage & 3D-Curb & NRS & Real Scene \\
\hline
Model Inference & 20.69 / 48.33 & 18.18 / 55.01 & 16.22 / 61.65 \\
Post-Processing & 26.96 / 37.09 & 26.08 / 38.34 & 19.02 / 52.58 \\
\hline
\end{tabular}
\label{Table7}
\vspace{-5pt}
\end{table}

\section{Limitation}
While the method presented in this paper demonstrates effective and accurate detection of curbs in road scenes, thus providing a basis for navigable area determination for autonomous driving, it currently has limitations in detecting curbs solely within the LiDAR point cloud. Due to factors such as the scanning angle, field of view, and obstructions inherent to LiDAR technology, some road areas remain undetected in the point cloud, leading to an inability of the model to extract corresponding curb features. This necessitates future research involving more advanced sensors to minimize scanning blind spots. Additionally, the development of a model incorporating a curb prediction module that operates effectively in areas without scanning blind spots is essential to mitigate the impact of these blind spots on curb detection.

\section{Conclusion}

In this paper, we established the 3D-Curb dataset, comprising 7,100 frames. To our knowledge, this is currently the largest and most diverse curb point cloud dataset with the most extensive range of annotated categories. Notably, this is also the first dataset to feature 3D point cloud annotations for curbs, which will significantly aid future related research. Within the CurbNet framework, we introduced the Multi-Scale and Channel Attention (MSCA) module, addressing the challenges of uneven distribution of curb features and the reliance on high-frequency z-axis features. 
Additionally, we introduce a novel adaptive loss function group to resolve the imbalance in the number of curb point clouds relative to other categories. Extensive experiments on both the NRS and 3D-Curb datasets demonstrated that our approach outperforms the current leading curb detection and point cloud segmentation models.
In the tolerance experiments, CurbNet achieved over 0.95 average performance in Precision, Recall, and F-1 score metrics at just 0.15m tolerance, setting a new standard. Furthermore, our post-processing approach of multi-clustering and curve fitting effectively eliminated noise in the curb results, enhancing the Precision, Recall, and F-1 score metrics to 0.8744, 0.8648, and 0.8696, respectively. Finally, the excellent detection performance and generalization of our proposed method were further verified in real scene experiments.

The 3D-Curb dataset and the CurbNet framework established in this study lay a foundation for future research in curb detection. In our upcoming research, we plan to create a more comprehensive dataset incorporating additional modalities. Similarly, we aim to explore and enhance the capabilities of the CurbNet framework, improving its performance in multi-modal data contexts.





{
\bibliographystyle{IEEEtran}
\bibliography{ref}
}


 


\begin{IEEEbiography}[{\includegraphics[width=1in,height=1.25in,clip,keepaspectratio]{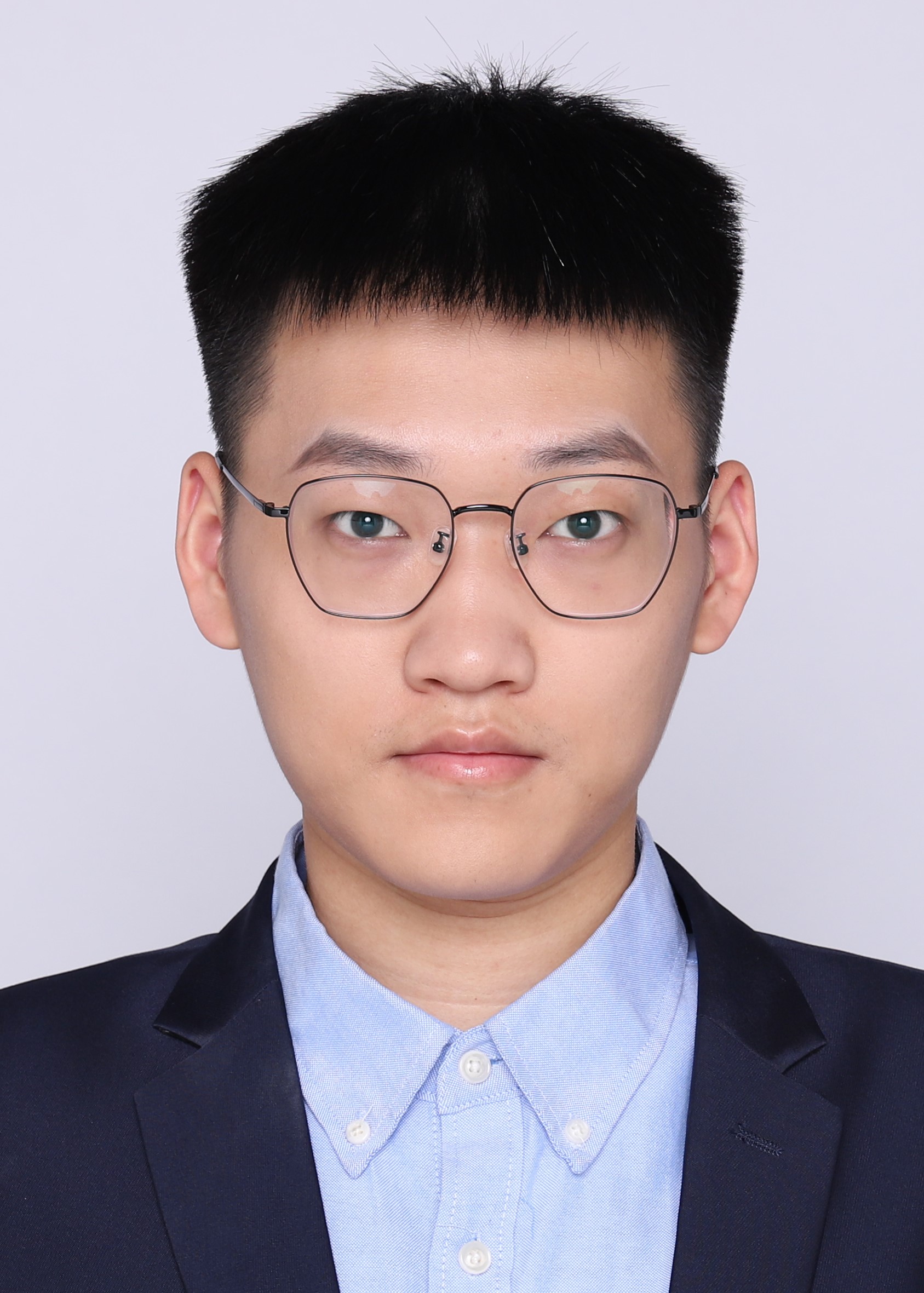}}]
{Guoyang Zhao} (Student Member, IEEE) received the B.Eng. degree in logistics engineering from Northeast Agricultural University, Harbin, China, in 2022, and the M.Phil. degree in robotics and autonomous systems from The Hong Kong University of Science and Technology (Guangzhou), Guangzhou, China, in 2024. He is currently pursuing the Ph.D. degree at the Intelligent Autonomous Driving Center, Robotics and Autonomous Systems Thrust, The Hong Kong University of Science and Technology, Guangzhou, China. His research interests include computer vision, robotics navigation, and deep learning.
\vspace{-25pt}
\end{IEEEbiography}

\begin{IEEEbiography}[{\includegraphics[width=1in,height=1.25in,clip,keepaspectratio]{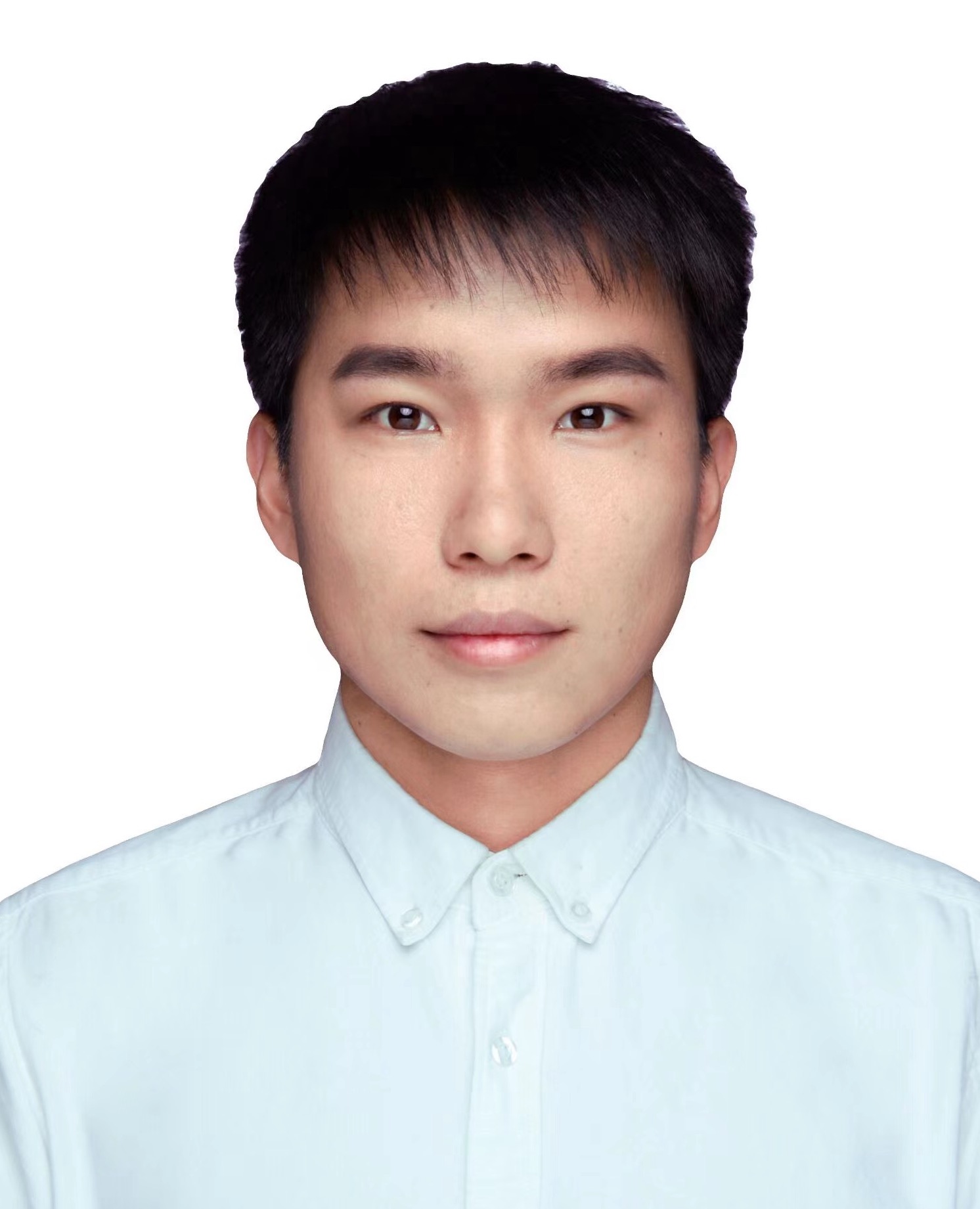}}]{Fulong Ma}  received the B.Eng. degree in automation from the University of Science and Technology of China, Hefei, China, in 2018. He is currently pursuing the Ph.D degree with the Robotics and Autonomous Systems Thrust, The Hong Kong University of Science and Technology (Guangzhou), Guangzhou, China.
His research interests include computer vision, sensor calibration, and deep learning.
\vspace{-25pt}
\end{IEEEbiography}

\begin{IEEEbiography}[{\includegraphics[width=1in,height=1.25in,clip,keepaspectratio]{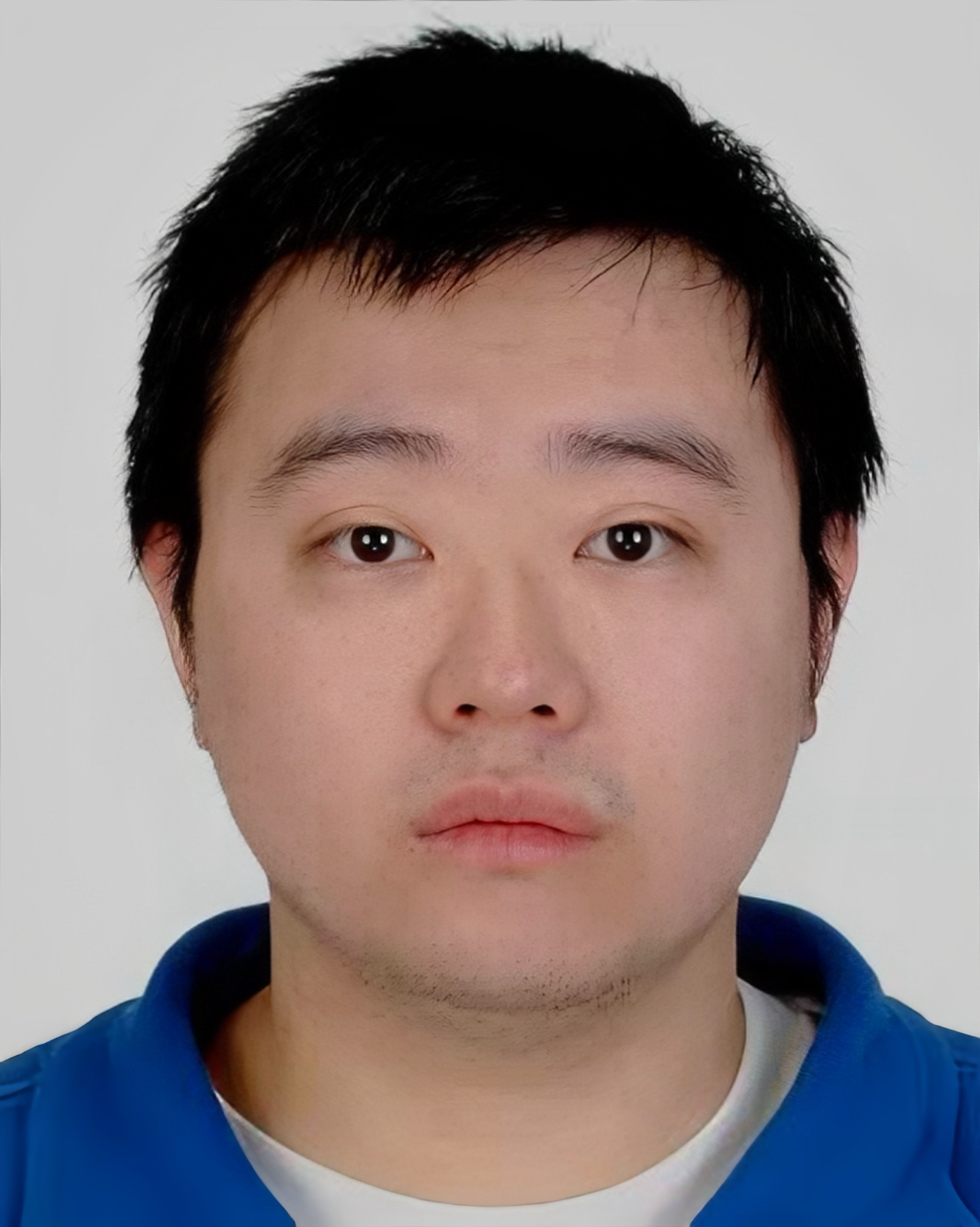}}]{Weiqing Qi} received the B.S. degree in Computer Science from University of California, Santa Barbara, CA, USA, in 2021, and the M.Phil. degree in robotics and autonomous systems from The Hong Kong University of Science and Technology (Guangzhou), Guangzhou, China, in 2024. His current research interests include lane detection, drivable area segmentation, and semantics segmentation, etc.
\vspace{-25pt}
\end{IEEEbiography}

\begin{IEEEbiography}[{\includegraphics[width=1in,height=1.25in,clip,keepaspectratio]{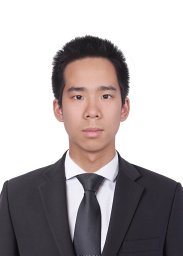}}]{Yuxuan Liu}  received the B.Eng. degree in Mechatronic from Zhejiang University, Zhejiang, China in 2019, and the Ph.D. degree in Electronic and Computer Engineering, The Hong Kong University of Science and Technology, Hong Kong, China, in 2023. His current research interests include autonomous driving, deep learning, robotics, visual 3D object detection, visual depth prediction, etc.
\vspace{-25pt}
\end{IEEEbiography}

\begin{IEEEbiography}[{\includegraphics[width=1in,height=1.25in,clip,keepaspectratio]{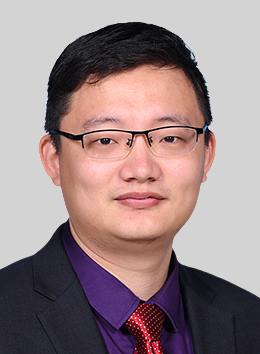}}]
{Ming Liu} received the B.A. degree in automation from Tongji University, Shanghai, China, in 2005, and the Ph.D. degree from the Department of Mechanical and Process Engineering, ETH Zurich, Zurich, Switzerland, in 2013, supervised by Prof. Roland Siegwart. During his master's study with Tongji University, he stayed one year with the Erlangen-Nunberg University and Fraunhofer Institute ISB, Erlangen, Germany, as a Visiting Scholar.
He is currently an Associate Professor with the Robotics and Autonomous Systems Thrust, The Hong Kong University of Science and Technology (Guangzhou), Guangzhou, China. He is also a founding member of Shanghai Swing Automation Ltd., Co. He is currently the Chairman of Shenzhen Unity Drive Inc., China. He has coordinated and been involved in NSF Projects and National 863-Hi-TechPlan Projects in China. From 2014 to 2015, He was an Assistant Professor with City University of Hong Kong, Hong Kong SAR, China. He was an Assistant Professor from 2017 to 2020 and an Associate Professor since 2020, with The Hong Kong University of Science and Technology, Hong Kong SAR, China.

He has published several papers in top journals including IEEE Transactions on Robotics and International Journal of Robotics Research. He was an Associate Editor for IEEE Robotics and Automation Letters, IET Cyber-Systems and Robotics, International Journal of Robotics and Automation, IEEE IROS Conference 2018, 2019 and 2020. He served as a Guest Editor of special issues in IEEE Transactions on Automation Science and Engineering. His research interests include dynamic environment modeling, deep learning for robotics, 3-D mapping, machine learning, and visual control. 
\end{IEEEbiography}

\begin{IEEEbiography}[{\includegraphics[width=1in,height=1.25in,clip,keepaspectratio]{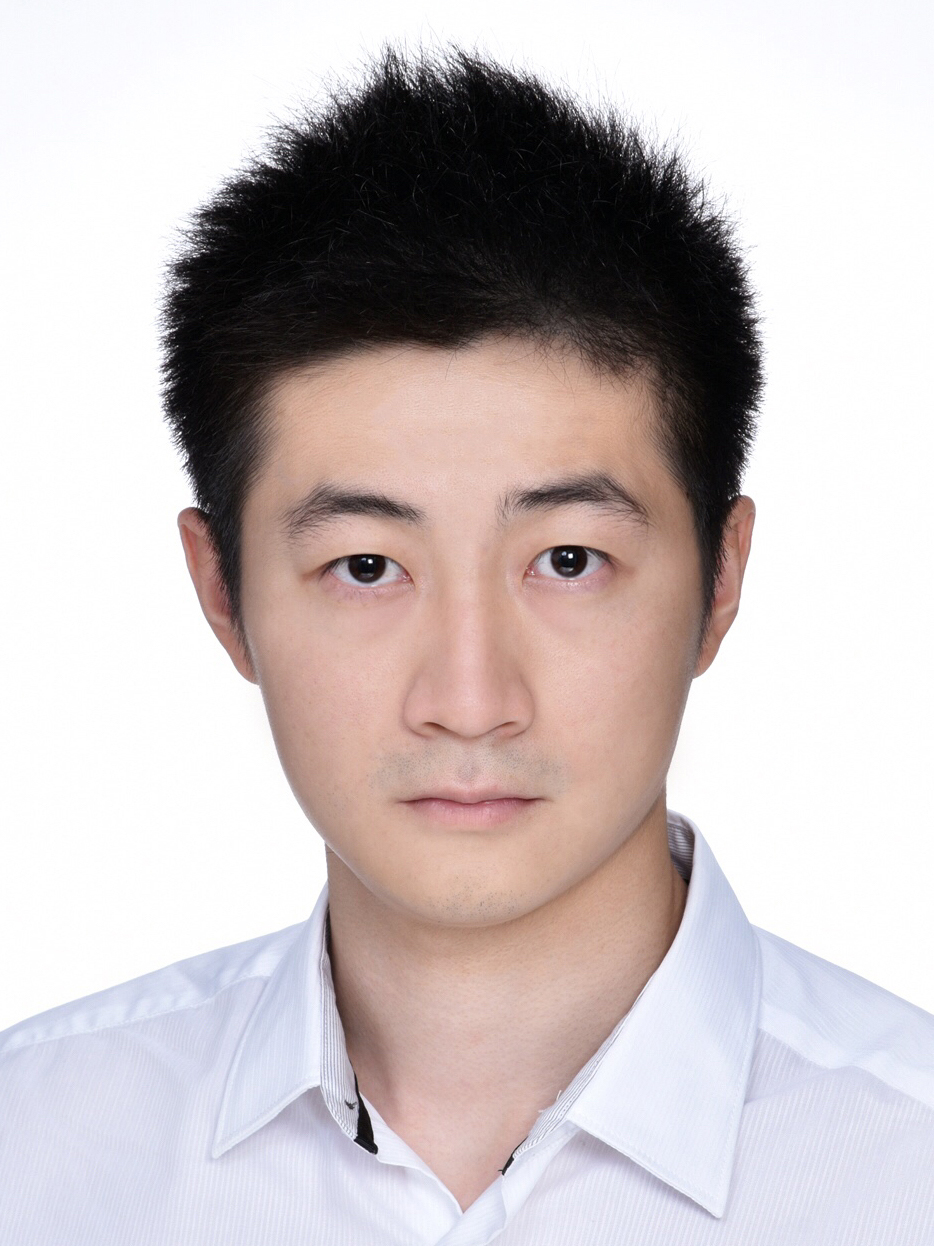}}]
{Jun Ma} (Senior Member, IEEE) received the B.Eng. degree with First Class Honours in electrical and electronic engineering from  Nanyang Technological University, Singapore, in 2014, and the Ph.D. degree in electrical and computer engineering from the National University of Singapore, Singapore, in 2018.
From 2018 to 2021, he held several positions at the National University of Singapore; University College London, London, U.K.; University of California, Berkeley, Berkeley, CA, USA; and Harvard University, Cambridge, MA, USA.  He is currently an Assistant Professor with the Robotics and Autonomous Systems Thrust, The Hong Kong University of Science and Technology (Guangzhou), Guangzhou, China, and also with the Division of Emerging Interdisciplinary Areas, The Hong Kong University of Science and Technology, Hong Kong SAR, China. He is also the Director of Intelligent Autonomous Driving Center, The Hong Kong University of Science and Technology (Guangzhou), Guangzhou, China.
His research interests include motion planning and control for robotics and autonomous driving.
\end{IEEEbiography}

\vspace{11pt}

\vfill

\end{document}